\documentclass[sigconf]{acmart}

\newcommand{\et}[2]{${#1}^{\pm{#2}}$}
\newcommand{\etb}[2]{$\mathbf{{#1}}^{\pm{#2}}$}
\newcommand{\DN}{HumanAct12}
\newcommand{\DNT}{PHSPD}
\usepackage{multirow}
\usepackage{multicol}
\usepackage{enumitem}

\copyrightyear{2020}
\acmYear{2020}
\setcopyright{acmcopyright}
\acmConference[MM '20]{Proceedings of the 28th ACM International Conference on Multimedia}{October 12--16, 2020}{Seattle, WA, USA}
\acmBooktitle{Proceedings of the 28th ACM International Conference on Multimedia (MM '20), October 12--16, 2020, Seattle, WA, USA}
\acmPrice{15.00}
\acmDOI{10.1145/3394171.3413635}
\acmISBN{978-1-4503-7988-5/20/10}

\begin{document}
\fancyhead{}

\title{Action2Motion: Conditioned Generation of 3D Human Motions}

\author{
Chuan Guo$^{1}$ , 
Xinxin Zuo$^{1,4}$, 
Sen Wang$^{1,4}$,
Shihao Zou$^{1}$, 
Qingyao Sun$^{2}$, 
Annan Deng$^{3}$ \and
Minglun Gong$^{4}$, 
Li Cheng$^{1}$}

\affiliation{
$^1$Department of Electrical and Computer Engineering, University of Alberta \\
$^2$Physical Sciences Division, University of Chicago \\
$^3$Graduate School of Arts and Sciences, Yale University \\
$^4$School of Computer Science, University of Guelph }


\email{{cguo2, lcheng5, szou2}@ualberta.ca, {xinxinzuo2353, wangsen}@gmail.com}









\begin{teaserfigure}
\centering
\includegraphics[width=0.87\textwidth]{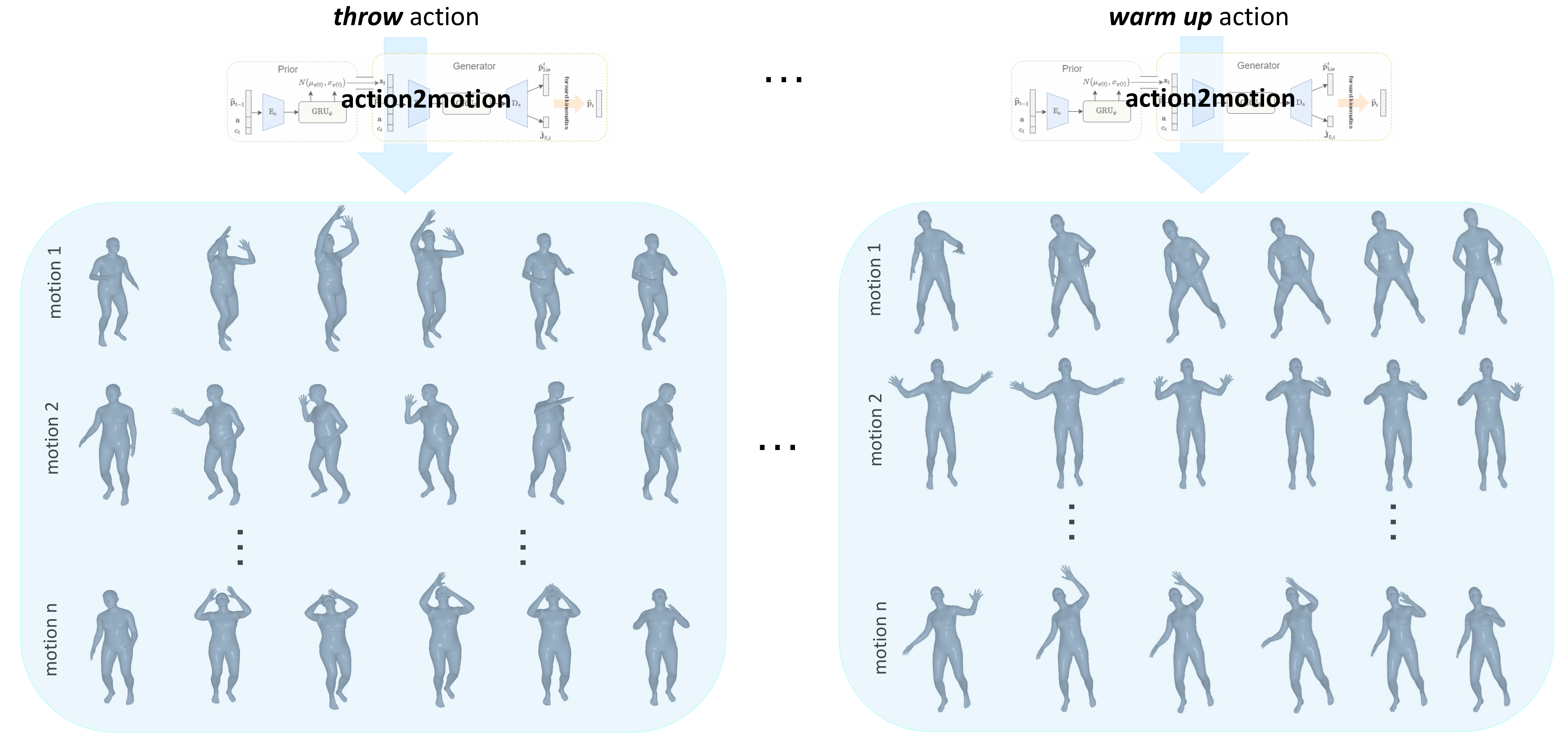}
\setlength{\abovecaptionskip}{0.1cm}
  \setlength{\belowcaptionskip}{-0.2cm} 
\caption{Conditioned on an action category (such as throw, warm up), our approach can generate a diverse set of natural 3D human motions. 
}
\label{fig:overview}
\vspace{2pt}
\end{teaserfigure}

\begin{abstract}
Action recognition is a relatively established task, where given an input sequence of human motion, the goal is to predict its action category. This paper, on the other hand, considers a relatively new problem, which could be thought of as an inverse of action recognition: given a prescribed action type, we aim to generate plausible human motion sequences in 3D. Importantly, the set of generated motions are expected to maintain its \textit{diversity} to be able to explore the entire action-conditioned motion space; meanwhile, each sampled sequence faithfully resembles a \textit{natural} human body articulation dynamics. Motivated by these objectives, we follow the physics law of human kinematics by adopting the Lie Algebra theory to represent the \textit{natural} human motions; we also propose a temporal Variational Auto-Encoder (VAE) that encourages a \textit{diverse} sampling of the motion space. 
A new 3D human motion dataset, \DN, is also constructed~\footnote{For more details, see our project website: https://ericguo5513.github.io/action-to-motion/}. Empirical experiments over three distinct human motion datasets (including ours) demonstrate the effectiveness of our approach. 
\end{abstract}




\maketitle

\section{Introduction}
Looking at human is often a central theme in the images and videos of our daily life. In recent years, we have evidenced a surge of interests and brilliant progresses in video synthesis and prediction of future frames~\cite{tulyakov2018mocogan,denton2018stochastic,suwajanakorn2015makes}. However, when coming to looking at human, it remains a significant challenge. This is, for example, evidenced in the efforts of many recent pixel-based video generation methods~\cite{yang2018pose,cai2018deep} -- the synthesized objects commonly possess strange appearances that deviate from photo-realistic human shapes, their motions are usually unsatisfactorily distorted. 
These observations pronounce the importance of properly representing human body poses and modeling their temporal articulations in automated video generation. It also inspires our investigation into the fundamental problem of action-conditioned generation of 3D human motions.


The problem of generating human motions is far from being trivial. The principal challenges are two folds: the generated motions should be sufficiently diverse to cover the broad range of ways individuals perform the same type of actions; each motion sequence is also expected to be visually realistic. 
Most existing efforts either work directly in 2D space, or require initial pose/motion. For instance, \cite{yang2018pose} generates 2D motions based on an action type and an initial pose in a deterministic way, and then synthesizes appearances in video frame by frame. In ~\cite{cai2018deep}, a pose generator and motion generator are trained progressively with the help of GANs for human 2D motion generation. Other methods~\cite{yan2018mt,kim2019unsupervised} directly generate 2D human motions by VAEs or GANs. However, direct modeling of motions in 2D is inherently insufficient to capture the underlying 3D human shape articulations. 
Existing methods~\cite{yang2018pose,cai2018deep,yan2018mt,lin20181} often approach human poses in terms of the joints' coordinate locations, which unnecessarily entangle the human skeletons and their motion trajectories, and introduce extra barriers in faithful modeling of human kinematics. 
To further complicate the matter, the variation of generated human dynamics could be severely limited by the initial priors~\cite{yang2018pose}. 

To address the aforementioned challenges, we consider to generate 3D human motions without prior conditions of initial poses or motions. 
Instead, a novel framework is proposed that consists of a conditional temporal VAE based on Lie algebra representation. 
Inspired by~\cite{denton2018stochastic}, we leverage the posterior distribution learned from previous poses as a learned prior to gauge the generation of present pose; by tapping into the RNN implementation, this learned prior also encapsulates temporal dependency across consecutive poses. 
The proposed modeling design thus facilitates our approach to go beyond the stereotypical motion sequences and uncover varying interpolations of the articulation trajectories in the training set.
For motion representation, the theory of Lie algebra has proven its effectiveness for articulate pose modeling in related tasks including pose estimation, action recognition, and motion prediction~\cite{liu2019towards,vemulapalli2014human,gui2018adversarial,huang2017deep}. Specifically, human pose is characterized as a kinematic tree based on physics principle of human full-body kinematics. There are multiple advantages of using Lie representation over the joint-coordinate representation: (i) Lie representation disentangles the skeleton anatomy, temporal dynamics and scale information; (ii) Lie representation faithfully encodes the anatomical constraints of skeletons, by following the physics law of forward kinematics; (iii) Lie algebra space is roughly an Euclidean space, thus the important linear algebra concepts deeply rooted in regression learning techniques could work again. As a by-product, the dimension of Lie algebra space naturally corresponds to the degrees of freedom (DoF), which is more compact comparing to joint coordinates. 
In practice, the adoption of Lie algebra notably mitigates the trembling phenomenon prevailing in joint coordinates representation, facilitates the generation of natural, lifelike motions, and accelerates the learning process. 
Empirically, training with Lie algebra requires only one tenth of the number of iterations by training with joint coordinates in reaching equilibrium.

To summarize, our key contributions are three-fold:
first, we aim to address a new problem of 3D motion sequence generation based on prescribed action categories. Our approach is to our knowledge the first in addressing such a problem;  
second, a novel Lie Algebra based VAE framework is proposed, capable of generating natural and diverse sets of human motions for prescribed action categories; 
and third, since the mainstream 3D human pose datasets are not directly applicable to our purpose, we curate own 3D human motion dataset. 
The NTU-RGB-D dataset~\cite{liu2019ntu} possesses large-scale 3D motions of various action categories; their pose annotation is severely inaccurate. We thus re-annotate this dataset with the help of a recent tool~\cite{kocabas2019vibe}. 


\section{Related Work}
\textbf{Multimodal 3D Human Dynamics Generation}: 
There have been several prior endeavors in generating 3D human dynamics from various modalities, including audio, text and images. 
As audio signals do not contain explicit information of pose structure and motion dynamics, a common strategy is to learn the correlation between two modalities, and to obtain intrinsic representation of poses and motions. In \cite{takeuchi2017speech}, upper body gestures are generated based on speech signals. \cite{tang2018dance} employs LSTM based autoencoder to capture music-to-dance mapping, and \cite{shlizerman2018audio} predicts body dynamics from violin and piano recital audios with LSTM model. A more recent work of \cite{lee2019dancing} generates human dancing dynamics conditioned on music in a non-deterministic manner. However, due to the difficulties of collecting data, most of them are conducted in 2D space.

Motions could also be inferred from text, an emerging subject that is more related to our problem. Pioneer efforts such as~\cite{lin20181,ahn2018text2action,plappert2018learning} mainly resort to encoder-decoder RNN architecture for language-to-pose translation. The work of \cite{ahuja2019language2pose} learns a joint embedding space between sentences and human pose sequences.  More recently, \cite{stoll2020text2sign} applies more sophisticated neural translation network equipped with GANs for text-to-sign prediction. 


The problem of action based human 2D motion generation is mostly related to our goal. Unfortunately there are few studies along this topic. Among them, \cite{cai2018deep} adopts two stage GAN framework to generate 2D human motion progressively. To our knowledge, our work is the first for action based 3D human motion generation.

\textbf{Skeleton-based Human Pose Representations}: 
Human pose representation plays a foundational role in modeling human motions. Proper representation may promote the realism of motions and foster performance robustness. Many existing methods directly utilize joints' coordinates to represent human skeletons~\cite{han2017space,hussein2013human}.  In ~\cite{wang2012mining}, pair-wise relative positions of joints are used to represent human skeleton, while in ~\cite{chaaraoui2014evolutionary}, uninformative keypoints are further pruned during modeling. Another branch of human pose representation is part-based, which considers a skeleton as a connected set of segments. In ~\cite{yacoob1999parameterized}, a human body is decomposed to five main parts, and a motion is parameterized by the temporal translations and rotations of body parts. \cite{gavrila1995towards} represents a human skeleton with 3D joint angles, whereas the temporal information is modeled using dynamic time warping ~\cite{muller2007information}. Finally, Lie group based methods such as \cite{vemulapalli2014human}, \cite{huang2017deep} and \cite{liu2019towards} characterize a human skeleton as kinematic chains, and represent 3D skeleton as a point in $\mathrm{SO}(3)$ or $\mathrm{SE}(3)$. 

\textbf{3D Human Motion Datasets}: 
There have been several datasets targeting at human actions, with pose or 3D joint positions being marked either by stationary kinematic sensors or manual labors. CMU MoCap~\cite{cmu2003mocap} and HDM05 ~\cite{muller2007mocap} contain more than 100,000 poses and 2000 motions or pose sequences organized into multiple categories. However, the distribution of sequences among categories is highly unbalanced; the categories are often not necessarily aligned with typical action types. UTKinect-Action ~\cite{xia2012view} and MSR-Action3D~\cite{li2010action} possess more clearly defined action annotations; they however contain much less number of motions. NTU-RGB-D~\cite{liu2019ntu} is so far the largest human motion dataset, consisting of over 100,000 motions belonging to 120 classes. Yet the pose annotations are from MS Kinect that are inexact and noisy; the motions are temporally unstable. We are therefore motivated to curate an in-house 3D human action dataset, \DN, as well as take effort in improving the pose annotation of NTU-RGB-D.

\section{Methodology}

\begin{figure*}[th]
  \vspace{-0.1cm}
  \centering
  \includegraphics[width=0.72\linewidth]{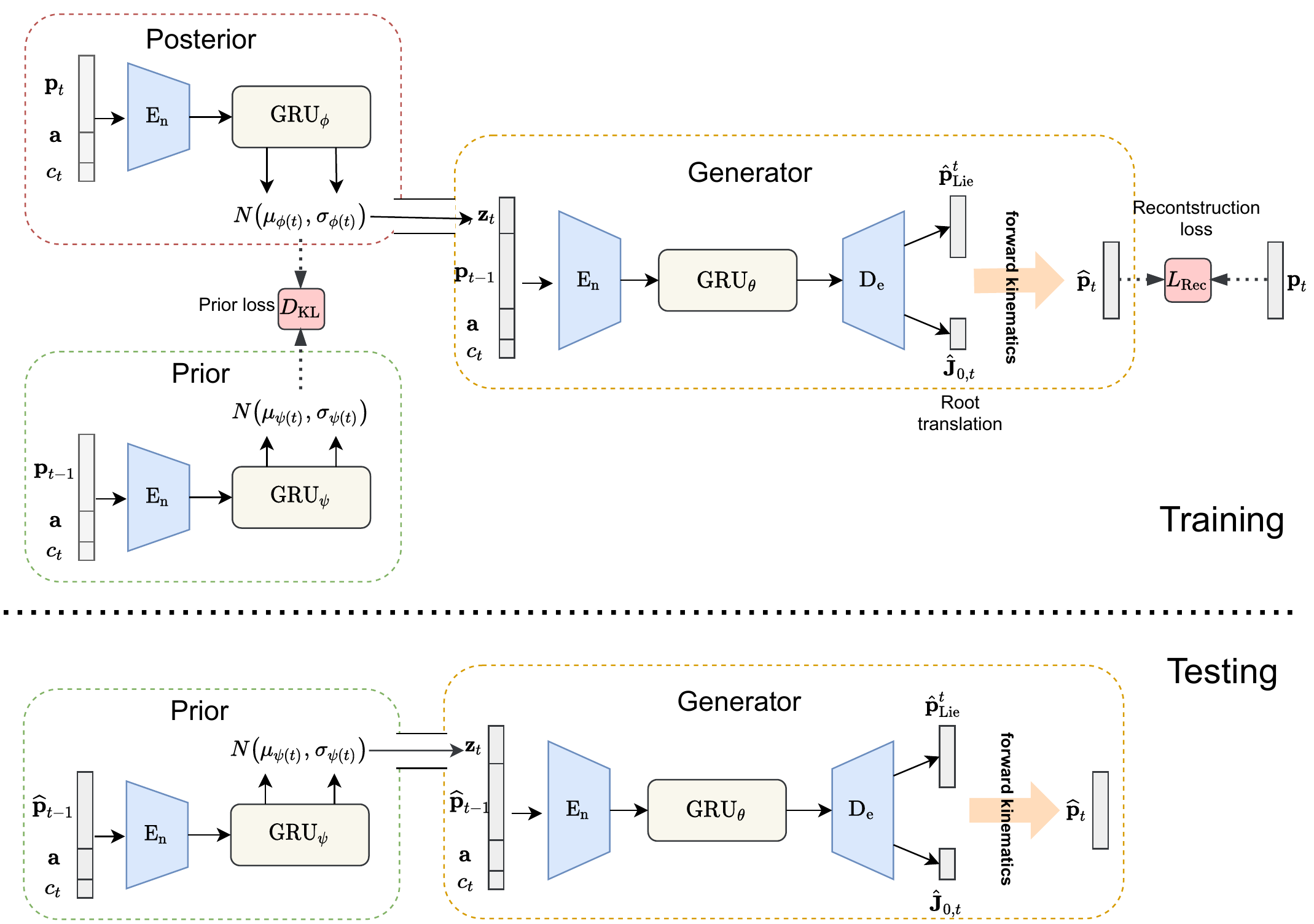}
  \caption{Overview of the proposed action2motion framework. (Top) Training phase: The input vector is a concatenation of action category $\mathbf{a}$, time count $c_t$, and pose vector($\mathbf{p}_t$ or $\mathbf{p}_{t-1}$). The prior is obtained from the partial sequence of poses so far, $\mathbf{p}_{1:t-1}$. KL-divergence is utilized to enforce posterior to be close to the prior. In generator, we first generate the Lie algebraic parameters and root translation of current pose, then produce the 3D joints positions by the physics law of forward kinematics (Sec\ref{subsec:lie_algebra}). (Bottom) Testing phase: A noise vector is sampled from the prior distribution, which kick-start the above-mentioned process in generate a 3D motion as sequence of poses.}
  \label{fig:architecture}
  \vspace{-0.3cm}
\end{figure*}

Given an action category $\mathbf{a}$, our goal is to generate a motion -- a 3D pose sequence, $M = [\mathbf{p}_1,...,\mathbf{p}_T]$, of length $T$. Here $\mathbf{p}_t \in \mathbb{R}^D$ is the human pose at time $t$. 
Our framework, illustrated in Figure~\ref{fig:architecture} and called \textbf{action2motion}, is essentially a conditional temporal VAE (Sec.\ref{subsec:ct_vae}) equipped with a Lie algebra pose representation (Sec.\ref{subsec:lie_algebra}). 

\subsection{Disentangled Lie Algebra Representation}
\label{subsec:lie_algebra}
A human pose could be depicted as a kinematic tree consisting of five kinematic chains: spine and four limbs. 
With the help of Lie algebra theory, we are able to distill skeleton anatomical information, motion trajectories and bone lengths from the alternative representation of 3D joint locations.

Let $E=\{e_1,...,e_N\}$ denote the set of oriented edges in a skeleton as oriented rigid body bones, with $N$ being the number of bones. 
Each of the bones $e_n$ is attached with its local coordinate system, with the bone itself orienting along the x-axis and the starting joint of the bone as origin. 
Let $e_n$ and $e_m$ be two consecutive bones along a kinematic chain, the coordinate transformation from $e_n$ to $e_m$ could be carried out through rotation and translation transformation matrices~\cite{xu2017lie,liu2019towards,huang2017deep}. Mathematically, a joint with coordinates $\mathbf{c}_n^i=(x_n^i, y_n^i, z_n^i)^\top$ w.r.t coordinate system $e_n$ will have coordinates $\mathbf{c}_m^i=(x_m^i, y_m^i, z_m^i)^\top$ w.r.t coordinate system $e_m$ with
\begin{equation}
    c_m^i = \left(
    \begin{array}{cc}
        R_{n} & \mathbf{d}_{n}  \\
        0 & 1
    \end{array}\right)\left(\begin{array}{c}
         \mathbf{c}_n^i  \\
          1
    \end{array}\right),
\end{equation}
where $R_{n} \in \mathbb{R}^{3\times3}$ is the rotation matrix, and $\mathbf{d}_{n} \in \mathbb{R}^3$ is the translation vector along x-axis. Furthermore, $\mathbf{d}_n$ could be written as $(b_n, 0, 0)^\top$ with $b_n$ denoting the bone length of $e_n$, a constant number over time.

Here, the rotation matrix $R_{n}$ between two local coordinates is an element of Special Orthogonal Group $\mathrm{SO}(3)$, which is a matrix Lie group. 
Hence, excluding bone lengths, the relative geometry between $e_n$ and $e_m$ is a point in $\mathrm{SO}(3)$ and the whole skeleton is represented as a point in $\mathrm{SO}(3)\times \mathrm{SO}(3)\times ... \times \mathrm{SO}(3)$, which is a matrix Lie group endowed with a differentiable manifold structure~\cite{liu2019towards,huang2017deep}. Similarly, the motion could be characterized as a curve in Manifold. 
To carry out optimization in Manifold, we could engage the proper mathematical apparatus of Lie algebra or tangent space that could be regarded as a flat space, thus our familiar linear algebra computations could be utilized.

\paragraph{Lie algebra $\mathfrak{so}(3)$} The 3 by 3 identity matrix $I_3$ is an element of a $\mathrm{SO}(3)$ and is referred to as the identity element in this group~\cite{murray1994mathematical}. The tangent space at the identity element $I_3$ of $\mathrm{SO}(3)$ is known as the Lie algebra space, $\mathfrak{so}(3)$, 
a 3-dimensional vector space spanned by the elements of a $3\times 3$ skew-symmetric matrix $\hat{W}$, as  
\begin{equation}
    \hat{W} = 
    \left(
    \begin{array}{ccc}
        0 & -w_3 & w_2  \\
        w_3 & 0 & -w_1  \\
        -w_2 & w_1 & 0
    \end{array}
    \right) .
\end{equation}
The association between a $R \in \mathrm{SO}(3)$ and its Lie algebra vector $ \mathbf{w} \in \mathfrak{so}(3)$ could be given by the logarithm map $\log_{\mathrm{SO}(3)}$: $ \mathrm{SO}(3)\rightarrow \mathfrak{so}(3)$, as
\begin{equation}
    \mathbf{w} =
    \left(
    \begin{array}{c}
        w_1  \\
        w_2  \\
        w_3
    \end{array}
    \right)
    = \frac{\theta}{2\sin{\theta}}
    \left(
    \begin{array}{c}
        R(3,2) - R(2,3) \\
        R(1,3) - R(3,1)  \\
        R(2,1) - R(1,2)
    \end{array}
    \right),
\end{equation}
where $\theta = \arccos{\frac{trace(R)-1}{2}}$ \cite{murray1994mathematical}. Since $\mathbf{w}$ is not uniquely mapped, we use the value with norm in range$[-\pi, \pi]$. Similarly, the inverse transformation is given by the exponential map $\exp^{\mathfrak{so}(3)}$: $ \mathfrak{so}(3) \rightarrow \mathrm{SO}(3)$. Please refer to \cite{murray1994mathematical} for more details.

\paragraph{Forward kinematics and motion trajectory} 
Let $K$ denote the number of kinematic chains, $m_k$ the number of joints in the $\mathrm{k}$-th chain, and $\mathbf{w}_i^k$ the Lie algebra parameter vector of joint $i$ in chain $k$. 
The Lie algebraic parameters of a pose could be formalized as a vector $\mathbf{p}_\mathrm{Lie}=({\mathbf{w}_1^1}^\top,...,{\mathbf{w}_{m_1}^1}^\top,...,{\mathbf{w}_1^K}^\top,...,{\mathbf{w}_{m_K}^K}^\top)^\top$. Then the 3D location of joint $\mathbf{J}_{i}^k$ in chain $k$ could be obtained by \textbf{forward kinematics},
\begin{equation}
\label{equ:forward}
         \mathbf{J}_{i}^k = \left[\prod_{j=1}^{i-1} \mathrm{exp}(\hat{W}_j^k) \right]\mathbf{d}_i^k + \mathbf{J}_{i-1}^k.
\end{equation}
Therefore, the Lie algebra parameters of a body pose could be transformed to 3D coordinates vector through joint-wise forward kinematics $\mathbf{\Gamma}(\mathbf{p}_\mathrm{Lie})$: $\mathbf{p}_\mathrm{Lie} \rightarrow \mathbf{p}$, where $\mathbf{p}=({\mathbf{J}_1^1}^\top,...,{\mathbf{J}_{m_1}^1}^\top,...,{\mathbf{J}_1^K}^\top,...,{\mathbf{J}_{m_K}^K}^\top)^\top$.
Forward kinematics typically starts from a root joint $\mathbf{J_0} \in \mathbb{R}^3$ denoting the spatial translation of the entire human body, 
which is independent from the pose (i.e. Lie algebra parameters). Consider a motion with $T$ consecutive poses, accordingly, the sequence $(\mathbf{J}_{0,1},...,\mathbf{J}_{0,T}) \in \mathbb{R}^{3\times T}$ forms the body motion trajectory, 
with $\mathbf{J}_{0,t}$ being the root translation of pose at time $t$.

Therefore, a human motion is a composition of three parts: motion trajectories, Lie algebraic parameters $\mathfrak{se}(3)$, and bone lengths. The Lie algebra of a motion is $M_\mathrm{Lie} = (\mathbf{p}_\mathrm{Lie}^1;...;\mathbf{p}_\mathrm{Lie}^T)$, with $T$ being the number of poses. 
Note the number of effective Lie algebra parameters $M_\mathrm{Lie}$ is noticeably less than the number of elements in matrix $M_\mathrm{Lie}$, 
since many bones rotate along less than three directions. 
At test stage, both Lie algebraic parameters $M_\mathrm{Lie}$ and motion trajectory 
are generated from our learned action2motion model. 
The skeletal bone lengths are acquired from typical real-life human bodies, which could be kept fixed during our motion generation process. 
The benefits are two-fold: (i) ensure the invariance of bone lengths in a motion, and (ii) enable the generation of motions with controllable body scale, i.e. by manually changing the bone length.

\subsection{Conditional Temporal VAE}
\label{subsec:ct_vae}

\subsubsection{Preliminaries of temporal VAE}

To generate a pose sequence, we would incorporate the Variational Auto-Encoder(VAE) unit into a recurrent module, with VAE handling the pose generation of each time-step, and RNN modeling the temporal dependence over time. 

Formally, given a real motion or pose sequence $\mathbf{M}=[\mathbf{p}_1,...,\mathbf{p}_T]$, VAE aims to maximize the probability of $\mathbf{M}$ sampled from the learned model distribution. 
At time $t$, RNN module $p_{\theta}(\mathbf{p}_t|\mathbf{p}_{1:t-1}, \mathbf{z}_{1:t})$ predicts the current pose $\mathbf{p}_t$ having latent variables $\mathbf{z}_{1:t}$, and conditioned on previous states $\mathbf{p}_{1:t-1}$. We rely on a variational neural network $q_\phi(\mathbf{z}_t|\mathbf{p}_{1:t})$ to approximate the true unknown posterior distribution $p_{\theta}(\mathbf{z}_t|\mathbf{p}_{1:t})$. This way, the objective of maximizing the data likelihood over the real sequence could be achieved as the following variational lower bound:
\begin{equation}
    \setlength{\abovedisplayskip}{1pt} 
    \setlength{\belowdisplayskip}{1pt} 
    \begin{aligned}
    \log  p_\theta(\mathbf{M}) &= \log \int_\mathbf{z} p_\theta(\mathbf{M}|\mathbf{z})p(\mathbf{z}) \\
            &\geq \mathbb{E}_{q_\phi(\mathbf{z}|\mathbf{M})}\log p_\theta(\mathbf{M}|\mathbf{z}) - D_{KL}(q_\phi(\mathbf{z}|\mathbf{M})\parallel p(\mathbf{z})) \\
            &= \sum_t \big[\mathbb{E}_{q_\phi(\mathbf{z}_t|\mathbf{p}_{1:t})}\log p_\theta(\mathbf{p}_t|\mathbf{p}_{1:t-1}, \mathbf{z}_{1:t}) \\ 
            &\,\,\,\,\,-D_{KL}\left(q_\phi(\mathbf{z}_t|\mathbf{p}_{1:t})\parallel p(\mathbf{z}_t)\right)\big].
    \end{aligned}
\end{equation}
Here, the first term of the lower bounding function encourages the generated sample to be sufficiently close to the real sample; 
the second term penalizes the KL-divergence between prior and posterior distributions. 

A simple form of the prior $p(\mathbf{z}_t)$ in VAE is a Gaussian with unitary variance, $\mathcal{N}(0, \mathbf{I})$. 
In practice, however, the prior distribution often vary with time. Take motions from the \textit{walk} category for example, 
sometimes the pose variance could be small (e.g. roaming); sometimes could be large (e.g. sudden change of directions or velocity). 
The prior thus needs to be flexible enough to accommodate these variations. 
Intuitively, the prior of present time could be guessed given the context of previous time steps. 
Following \cite{denton2018stochastic}, we parameterize the prior with a variational neural network $p_\psi(\mathbf{z}_t|\mathbf{p}_{1:t-1})$ conditioned on previous steps $\mathbf{p}_{1:t-1}$. 
Therefore, the variational lower bound of the sequence could be re-written as
\begin{equation}
    \begin{aligned}
    \log  p_\theta(\mathbf{M}) \geq 
    &\sum_t \bigg[\mathbb{E}_{q_\phi \left(\mathbf{z}_t|\mathbf{p}_{1:t} \right)} \log p_{\theta} \left( \mathbf{p}_t|\mathbf{p}_{1:t-1}, \mathbf{z}_{1:t} \right) \\ 
    &-D_{KL} \left( q_\phi \left(\mathbf{z}_t|\mathbf{p}_{1:t} \right)\parallel p_\psi \left(\mathbf{z}_t|p_{1:t-1} \right) \right) \bigg].
    \end{aligned}
\end{equation}
The constraint between prior and posterior distribution further encourages temporal consistency. 

\subsubsection{Our Approach}
Figure~\ref{fig:architecture} depicts the architecture of our framework, which has three components: posterior network, prior network, and generator. 
The action category input is represented as an one hot vector $\mathbf{a}$. 
In addition, a time counter $c_t\in [0,1]$ is specifically used to keep record of where we are in the sequence generation progress, calculated as $\frac{t}{T}$. 
The rest of the input vector contains as a subvector the current pose $\mathbf{p}_t$. 
All encoders $\mathrm{E_n}$ and decoders $\mathrm{D_e}$ are modeled as linear fully connected layers. Prior network ($p_\psi$) and posterior ($q_\phi$) share the same structure, 
but with different parameter values of the GRUs 
and encoders. 
The generator, $p_\theta$, first produces the Lie algebraic parameters $\hat{\mathbf{p}}_{\mathrm{Lie}, t}$ and root translation $\hat{\mathbf{J}}_{0,t}$ via linear layers and $\mathrm{GRU}_{\theta}$, then obtains the 3D positions of skeleton $\hat{\mathbf{p}}_t$ via the forward kinematics of Eq.\eqref{equ:forward}. 
In training phase, the pose generation process at time step $t$ is
\begin{equation}
    \begin{aligned}
    \mathbf{h}_t &= \mathrm{E_n}(\mathbf{p}_t, \mathbf{a}, c_t),\,\,\, c_t = \frac{t}{T} \\
    \left( \mu_\phi(t), \sigma_\phi(t) \right) &= \mathrm{GRU}_\phi(\mathbf{h}_t)\\
    \mathbf{z}_t &\sim \mathcal{N}(\mu_\phi(t), \sigma_\phi(t)) \\
    \mathbf{g}_t &= \mathrm{E_n}(\mathbf{p}_{t-1}, \mathbf{a}, c_t, \mathbf{z}_t) \\
    \mathbf{l}_t &= \mathrm{GRU}_\theta(\mathbf{g}_t)\\
    \left( \hat{\mathbf{p}}_{\mathrm{Lie}, t}, \hat{\mathbf{J}}_{0,t} \right) &= \mathrm{D_e}(\mathbf{l}_t) \\ 
    \hat{\mathbf{p}}_t &= \mathbf{\Gamma}(\hat{\mathbf{p}}_{\mathrm{Lie}, t}, \hat{\mathbf{J}}_{0,t})
    \end{aligned}
\end{equation}

Similarly, the prior distribution $p_\psi(\mathbf{z}_t|\mathbf{p}_{1:t-1}, \mathbf{a}, c_t)$ is formed as follows, 
\begin{equation}
    \begin{aligned}
    \mathbf{h}_{t-1} &= \mathrm{E_n}(\mathbf{p}_{t-1}, \mathbf{a}, c_t),\,\,\, c_t = \frac{t}{T} \\
    \left( \mu_\psi(t), \sigma_\psi(t) \right) &= \mathrm{GRU}_\psi(\mathbf{h}_{t-1}). \\
    \end{aligned}
\end{equation}

In testing phase, as real data $\mathbf{p}_t$ is not available, we instead sample $\mathbf{z}_t$ from the prior distribution $p_\psi({z}_t|\cdot)$ to generate $\hat{\mathbf{p}}_t$.

\subsubsection{A mixed training strategy}
As shown in Figure~\ref{fig:architecture}, in testing phase, output from the previous time step $\hat{\mathbf{p}}_{t-1}$ is used as input to generate current pose, $\hat{\mathbf{p}}_t$; while in training phase, ground truth data $\mathbf{p}_{t-1}$ is employed in producing $\hat{\mathbf{p}}_{t}$, a technique known as \textit{teacher forcing} in sequence prediction. However, the discrepancy of curricula between training and testing compromises the stability of the trained model. An alternative approach is to apply the last generated pose $\hat{\mathbf{p}}_{t-1}$ as input into current step in training, which also raises concern on divergence: as soon as the prediction $\hat{\mathbf{p}}_{t}$ deviate from ground truth $\mathbf{p}_t$, the error may escalate in follow-up steps $t+1:T$.
In our context, a mixed strategy is adopted by randomly choosing among these two scenarios from a Bernoulli distribution $V \sim \mathrm{Bernoulli}(p_{\mathrm{tf}})$. Specifically, \textit{teacher forcing} is chosen for the entire sequence $\mathbf{p}_{1:T}$ if $V$ is 1, and vice versa.
As a boundary condition, when creating the initial pose $\hat{\mathbf{p}}_1$, its previous pose input $\mathbf{p}_0$ for prior $q_\psi$ is a zero vector. 

\subsubsection{Final Objective}
Put together, our final objective function becomes
\begin{equation}
    \begin{aligned}
    \mathcal{L}_{\theta,\phi,\psi} &= -\sum_{t=1}^T \bigg[ \mathbb{E}_{q_\phi(\mathbf{z}_t|\mathbf{p}_{1:t}, \mathbf{a}, c_t)}\log p_\theta(\mathbf{p}_t|\mathbf{p}_{1:t-1}, \mathbf{z}_{1:t}, \mathbf{a}, c_t) \\ &\,\,\,\,\,-\lambda D_{KL} \left(q_\phi \left(\mathbf{z}_t|\mathbf{p}_{1:t}, \mathbf{a}, c_t \right)\parallel p_\psi \left(\mathbf{z}_t|\mathbf{z}_{1:t-1}, \mathbf{a}, c_t \right)\right) \bigg],
    \end{aligned}
\end{equation}
where $\lambda$ is a tuning parameter for trade-off between reconstruction error and KL-divergence. 
In our implementation, the reconstruction term reduces to an $\ell_2$ penalty between $\hat{\mathbf{p}}_t$ and $\mathbf{p}_t$; 
the model is also trained with the re-parameterization trick~\cite{kingma2013auto}.

\section{Experiments}

\begin{figure*}[t]
  \centering
  \includegraphics[width=0.99\linewidth]{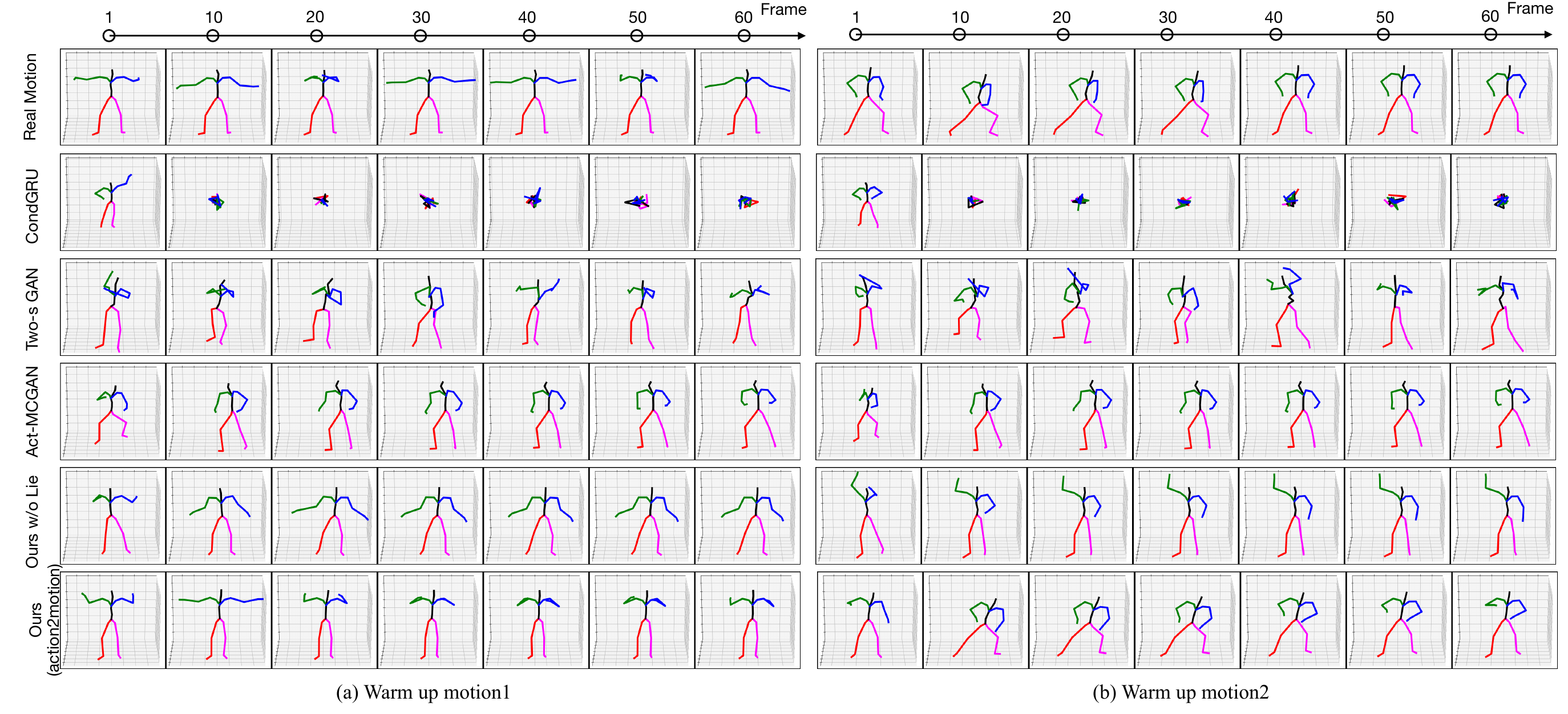}
  \setlength{\abovecaptionskip}{0.02cm}
  \setlength{\belowcaptionskip}{-0.2cm} 
  \caption{Given action \textit{warm up}, two motions are sampled from each motion source. In CondGRU, the first frame is required as part of the input; Nevertheless, in the remaining frames, all the joints collapse toward the root joints. Two-stage GAN, on the other hand, tends to produce jerking motions; the generated poses are also prone to perceivable distortion. Act-MoCoGAN oftentimes produce closely resembled pose sequences; its generated sequence also quickly reduce to stationary poses. Ours w/o Lie is able to generate motions with variation; meanwhile, its pose results often contain visible defects such as varying bone lengths and unnatural skeletons. Compared to the baselines, our approach (actoin2motion) excels in generating multiple distinct and realistic motions.}
  \label{fig:quanlitative_result1}
  \vspace{-0.3cm}
\end{figure*}

\subsection{Datasets}
The dataset of 3D human motions considered in our context is to possess multiple distinct action categories, with each action type containing a considerable amount of motions of diverse styles, and with proper pose annotations. Unfortunately the mainstreaming datasets are not directly applicable. This has led us to revamp two existing datasets, NTU-RGB-D~\cite{liu2019ntu} and CMU MoCap with proper re-annotations, as well as constructing our in-house dataset, \DN. 
Note in these datasets, the skeletons are all composed of 5 kinematic chains, and the root joint is located at pelvis.


\textbf{NTU-RGB-D~\cite{liu2019ntu}} originally contains 120 action types of 106 subjects. Its pose representation (3D joint positions) is from MS Kinect readout, which is known to unreliable and temporally inconsistent. We apply a state-of-art method~\cite{kocabas2019vibe} to re-estimate the 3D positions of 18 body joints (i.e. 19 bones) from the point cloud formed by aligning synchronized video feeds from multiple cameras. Note the poses are not necessarily matched perfectly to their true poses. It is sufficient here to be perceptually natural and realistic. A subset of 13 distinct actions are further chosen in our empirical evaluation, including e.g. \textit{cheer up, pick up, salute}, constituting 3,900 motion clips. 

\textbf{CMU MoCap} originally consists of 2,605 motion sequences, which however is not categorized by action type. Based on their motion descriptions, we identify 8 disparate actions, including \textit{running, walking, jumping, climbing}, and manually re-organize 1,088 motions. Here each skeleton is annotated with 22 3D joints (19 bones). In practice, the pose sequences are down-sampled to a frequency of 12 HZ from 100 HZ.

\textbf{\DN} is the in-house dataset which is adopted from an existing dataset PHSPD~\cite{polardataset,zou2020detailed}, consisting of 1,191 motion clips and 90,099 frames in total, with hierarchical action type annotations. To be specific, all motions are organized into 12 action categories, including \textit{warm up}, \textit{lift dumbbell}, and 34 subcategories including \textit{warm up eblowback}, \textit{lift dumbbell with right hand}). The fine-grained annotations offer more delicate details of the motions. In our experiments, we only use the coarse-grained action annotations. Compared to NTU-RGB-D, the 3D position annotations are more accurate, and the pose sequences are more stable. Compared to CMU MoCap, our \DN has more organized action annotation, with more balanced number of motions per action. 
A body pose contains 24 joints (23 bones).

\subsection{Experiment Results}

\subsubsection{Evaluation Metrics}
Our quantitative evaluation is to examine the \textbf{naturality} and \textbf{diversity} of the generated 3D motions. Following \cite{lee2019dancing}, four metrics are considered for a comprehensive evaluation of the results: recognition accuracy, Frechet Inception Distance (FID)~\cite{heusel2017gans}, diversity, and multimodality. 
FID is the most important indicator in our experiments: A \textbf{lower} FID suggests a better result. Note a result is claimed better than others on diversity and multimodality, only if its diversity and multimodality scores are \textbf{closer} to their respective values obtained from real motions. 
Finally, since there is no standard motion feature extractor, we train a standard RNN action recognition classifier for each dataset, and use its final layer as the motion feature extractor. 
Below are details of the four metrics:
\begin{itemize}[leftmargin=*]
    \item \textbf{Frechet Inception Distance} (FID): Features are extracted from 3,000 generated motions and real motions (obtained by sampling with replacement from the test set). Then FID is calculated between the feature distribution of generated motions vs. that of the real motions. FID is an important metric widely used to evaluate the overall quality of generated motions.
    \item \textbf{Recognition Accuracy}: We use a pre-trained RNN action recognition classifier to classify the 3,000 motions, and calculate the overall recognition accuracy. The recognition accuracy indicates the correlation of the motion and its action type.
    \item \textbf{Diversity}: Diversity measures the variance of the generated motions across all action categories. From a set of all generated motions from various action types, two subsets of the same size $S_d$ are randomly sampled. Their respective sets of motion feature vectors $\{\mathbf{v}_1,...,\mathbf{v}_{S_d}\}$ and $\{\mathbf{v}_1',...,\mathbf{v}_{S_{d}'}\}$ are extracted. The diversity of this set of motions is defined as
    \begin{equation}
        \mathrm{Diversity} = \frac{1}{S_d}\sum_{i=1}^{S_d}\parallel \mathbf{v}_i-\mathbf{v}_i' \parallel_2.
    \end{equation}
    $S_d=200$ is used in experiments.
    
    \item \textbf{Multimoldality}: Different from diversity, multimodality measures how much the generated motions diversify within each action type. Given a set of motions with $C$ action types. For $c$-th action, we randomly sample two subsets with same size $S_l$, and then extract two subset of feature vectors $\{\mathbf{v}_{c,1},...$ $,\mathbf{v}_{c,S_l}\}$ and $\{\mathbf{v}_{c,1}',...,\mathbf{v}_{c,S_l}'\}$. The multimodality of this motion set is formalized as
    \begin{equation}
        \mathrm{Multimodality} = \frac{1}{C \times S_l} \sum_{c=1}^C \sum_{i=1}^{S_l} \left\| \mathbf{v}_{c,i}-\mathbf{v}'_{c,i} \right\|_2.
    \end{equation}
    ${S_l}=20$ is used in experiments.
\end{itemize}

\begin{table*}[t]
  \caption{Performance evaluation on \DN and NTU-RGB-D.($\pm$ indicates 95\% confidence interval, and $\rightarrow$ means the closer to Real motions the better.)}
  \vspace{-5pt}
  \label{tab:performance1}
  \begin{tabular}{l c c c c c c c c c c c}
    \toprule
    \multirow{2}{*}{Methods} & \multicolumn{4}{c}{\DN} & & \multicolumn{4}{c}{NTU-RGB-D} \\
    \cline{2-5}
    \cline{7-10}
                    & FID$\downarrow$ & Accuracy$\uparrow$ & Diversity$\rightarrow$& Multimodality$\rightarrow$ &  & FID$\downarrow$ & Accuracy$\uparrow$ & Diversity$\rightarrow$ & Multimodality$\rightarrow$\\   
    \midrule

            \textbf{Real motions}    & \et{0.092}{.007}  &  \et{0.997}{.001}& \et{6.853}{.053}& \et{2.449}{.038}  & & \et{0.031}{.004}  &  \et{0.999}{.001}& \et{7.108}{.048}& \et{2.194}{.025}\\
    \midrule
            CondGRU   & \et{40.61}{.144}  &  \et{0.080}{.002}& \et{2.381}{.020}& \etb{2.341}{.036}  & & \et{28.31}{.138}  &  \et{0.078}{.001}& \et{3.663}{.024}& \et{3.578}{.027} \\
            Two-stage GAN    & \et{10.48}{.089} &  \et{0.421}{.006}& \et{5.960}{.049}& \et{2.805}{.036}  & & \et{13.86}{.091}  &  \et{0.202}{.003}& \et{5.328}{.039}& \et{3.490}{.027} \\
            Act-MoCoGAN   & \et{5.610}{.113}  &  \et{0.793}{.004}& \etb{6.752}{.071}& \et{1.055}{.017}  & & \et{2.723}{.019}  &  \etb{0.997}{.001}& \et{6.920}{.061}& \et{0.907}{.009}\\
    \midrule
            Ours w/o Lie    & \et{3.299}{.079}  &  \et{0.656}{.005}& \et{6.742}{.046}& \et{4.248}{.037}  & & \et{0.540}{.047}  &  \et{0.832}{.004}& \et{6.926}{.049}& \et{3.443}{.052}\\
            Ours    & \etb{2.458}{.079}  &  \etb{0.923}{.002}& \et{7.032}{.038}& \et{2.870}{.037}  & & \etb{0.330}{.008} &  \et{0.949}{.001}& \etb{7.065}{.043}& \etb{2.052}{.030}\\
    \bottomrule
  \end{tabular}
  \vspace{-0.3cm}
\end{table*}

\begin{table}[ht]
  \caption{Performance evaluation on CMU MoCap Dataset.}
  \label{tab:performance2}
  \begin{tabular}{l c c c c}
    \toprule
    \multirow{2}{*}{Methods} & \multicolumn{4}{c}{CMU MoCap} \\
    \cline{2-5}
                    &FID$\downarrow$ & Acc$\uparrow$ & Div & MModality \\   
    \midrule

            \textbf{Real motions}    & \et{0.065}{.006} & \et{0.93}{.002} & \et{6.13}{.079} & \et{2.72}{.066}\\
    \midrule
            CondGRU    & \et{51.72}{.123} & \et{0.09}{.001} & \et{0.79}{.011} & \et{0.75}{.016} \\
            Two-stage GAN & \et{14.34}{.107} & \et{0.17}{.003} & \et{4.41}{.064} & \etb{1.62}{.024} \\
            Act-MoCoGAN    & \et{11.15}{.074} & \et{0.44}{.005} & \et{5.28}{.069} & \et{1.51}{.022}  \\
    \midrule
            Ours w/o Lie    & \et{2.994}{.052} & \et{0.37}{.004} & \et{5.79}{.044} & \et{5.00}{.045}  \\
            Ours    & \etb{2.885}{.116} & \etb{0.68}{.003} & \etb{6.50}{.061} & \et{4.12}{.056} \\
    \bottomrule
  \end{tabular}
  \vspace{-0.2cm}
\end{table}

\subsubsection{Comparison Methods}
The problem of action conditioned 3D human motion generation is new. As a result there are few existing methods to compare with. 
Here, the following state-of-art methods from related areas are adapted for fair comparisons:
\begin{itemize}[leftmargin=*]
    \item \textbf{CondGRU}. We use conditional GRU as our deterministic baseline. Vanilla RNN model structure is considered for audio-to-motion translation in \cite{shlizerman2018audio}. Here, we make minor modification that the model takes the condition vector and pose vector together as input at current step and output the pose vector for next step.
    \item \textbf{Two-stage GAN}. A two-stage GAN is proposed in \cite{cai2018deep} for action conditioned 2D human motion generation. A pose generator is first trained by WGAN, which is then plugged into a motion generator. The motion generator is learned to yield noise vector for pose generator at each time step, and a motion discriminator is employed to enforce temporal smoothness. Necessary modifications are made to this method to work in 3D. 
    \item \textbf{Act-MoCoGAN}. MoCoGAN~\cite{tulyakov2018mocogan} is proposed for video generation, which produces a sequence of video frames from noise vectors and certain contents. By keeping the original architecture, we amend the video and image discriminators to motion and pose discriminators, respectively, for human dynamics generation.
    \item \textbf{Ours w/o Lie}. Here the Lie algebra representation in our action2motion approach is removed such that, the generator directly outputs the 3D position of joints. It is used to evaluate the effect of the proposed Lie algebra representation.
\end{itemize}

\subsubsection{Qualitative Evaluation}

Figure~\ref{fig:quanlitative_result1} compares the visual results of motions generated from different methods based on same action types. For conditional GRU, generated poses all collapse to a set a spatial points near the root joint, which shows the inefficacy of simple RNN models toward non-deterministic generative task. It is worth noting that motion generation should not be a one-to-one mapping process. Instead, the generated motions are expected to be \textit{close} to the real motions in terms of their respective distributions. 

We first investigate these methods on the motion naturality and correlations to prescribed action type. Two-stage GAN~\cite{cai2018deep} generates roughly human-like poses; on the other hand, temporally the pose sequence typically contains jerking unnatural movements. Act-MoCoGAN can generate smoother motions with reasonable human poses; however, it tends to quickly descend into frozen poses even in a relatively short motion sequence. 
This may be partially attributed to the observation that learning sequential dependency across poses may go beyond the capability of standard GAN models. 
Ours w/o Lie algebra produces more natural motions that are also reasonable w.r.t. the given action type; there is still a notable gap though, when comparing to real motions. For instance, in the left sequence of Fig.~\ref{fig:quanlitative_result1}, the human hands (i.e. blue and green lines) become longer from $t=1$ to $t=20$, and suddenly shrink in their lengths from $t=20$ to $t=60$. Consequently, the skeleton articulation is somewhat unnatural. Moreover, the latter portion of generated motions gradually freeze into fixed poses. The observed artifacts may be attributed to the nature of 3D coordinate representation that comes with strong entanglement and weak anatomical constraints. In contrast, our approach is capable of generating visually appealing motions that are also sensible to the given action type. 

Diversity and multimodality are also important evaluation criteria. As observed from Figure~\ref{fig:quanlitative_result1}, motions from Two-stage GAN are not well recognizable; Act-MoCoGAN suffers from low variations, which may due to the \textit{mode collapse} issue of typical GAN methods. In contrast, our action2motion is able to produce diverse motions. It highlights the merit of the proposed temporal conditional VAE on maintaining variations in sequential scenarios.

\subsubsection{Quantitative Comparison}

Evaluation results on \DN\ and NTU-RGB-D datasets are presented in Table~\ref{tab:performance1}, and the results over CMU MoCap dataset are shown in Table~\ref{tab:performance2}. For fair comparison, each experiment is repeated 20 times, and a statistical interval with 95\% confidence is reported. 
Among the four metrics, FID is the most important indicator in evaluating the overall performance of a model. 
Meanwhile, recognition accuracy reflects the quality of models and the correlation with action type. 
%
%
%
Overall we have the following observations from the comparisons in Table~\ref{tab:performance1} and Table~\ref{tab:performance2} over four metrics and across three datasets.
First, our approach, action2motion, clearly outperforms the comparison methods on FID. 
Deterministic methods such as CondGRU~\cite{shlizerman2018audio} is incapable of handling with such one-to-many generative task. 
The GANs employed in both two-stage GAN~\cite{cai2018deep} and Act-MoCoGAN~\cite{tulyakov2018mocogan} help mitigating the problem, yet still not well enough. 
%
%
For recognition accuracy, our action2motion yields highest accuracy over \DN and CMU MoCap dataset, which suggests the potential to generate highly recognizable motions and to capture the characteristics of action types. Act-MoCoGAN gains comparable accuracy. 
The explanation may be traced to the fact that Act-MoCoGAN has already incorporated the action classification loss in its training procedure. Meanwhile our action2motion reaches comparable or higher performance even without such explicit constraints during training; moreover, the use of Lie algebra representation significantly improves the recognition accuracy of generated motions over three datasets. The relative poor performances on CMU MoCap dataset may be attributed to high complexity of motions in this dataset.

The metrics of diversity and multimodality are important metrics complementary to FID and recognition accuracy. 
Note that for diversity and multimodality, the higher values are not necessarily better; instead the values are best to be close to those obtained from the set of real motions. 
Overall action2motion attains the closest values to real motions' for diversity and multimodality; 
Interestingly, compared to ours w/o Lie, introduce of Lie algebra representation lowers the abnormally high multimodality score while facilitating a proper level of diversity in all experiments. This, we believe, may come from the strength of action2motion in following the physics law, where Lie algebra representation enforces the production of valid motions from the motion kinematics manifold. 

\begin{figure}[t]
  \centering
  \includegraphics[width=0.85\linewidth]{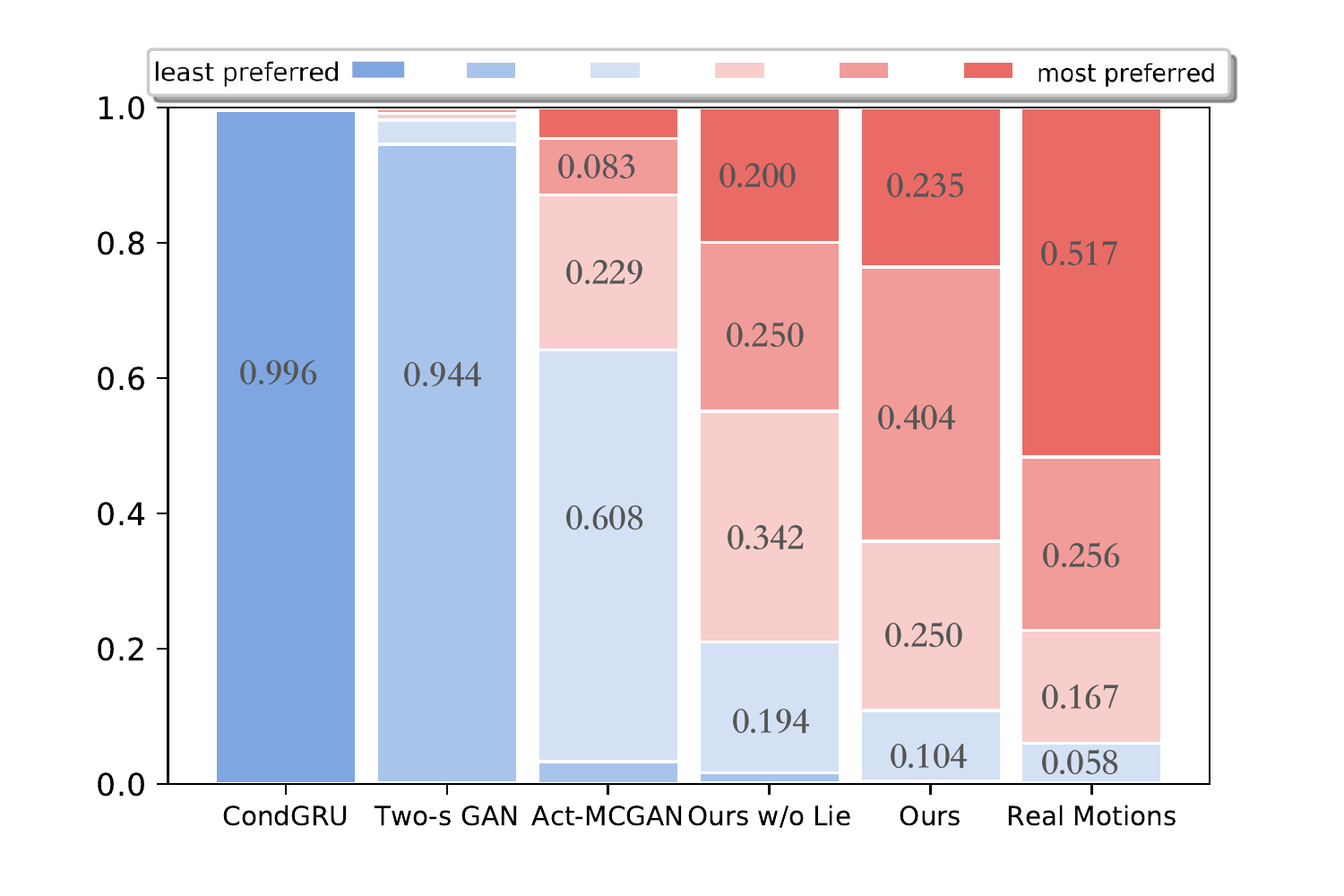}
  \setlength{\abovecaptionskip}{0.1cm}
  \setlength{\belowcaptionskip}{-0.2cm}  
  \vspace{-7pt}
  \caption{Preference results of generated motions. Bars with different color indicates the percentage of corresponding preference of each source. For example, darkest blue bar indicates the percentage of motions ranked as least preferred in each source.}
  \label{fig:rank_hist}
  \vspace{-0.3cm}
\end{figure}

\begin{figure}[t]
  \centering
  \includegraphics[width=0.725\linewidth]{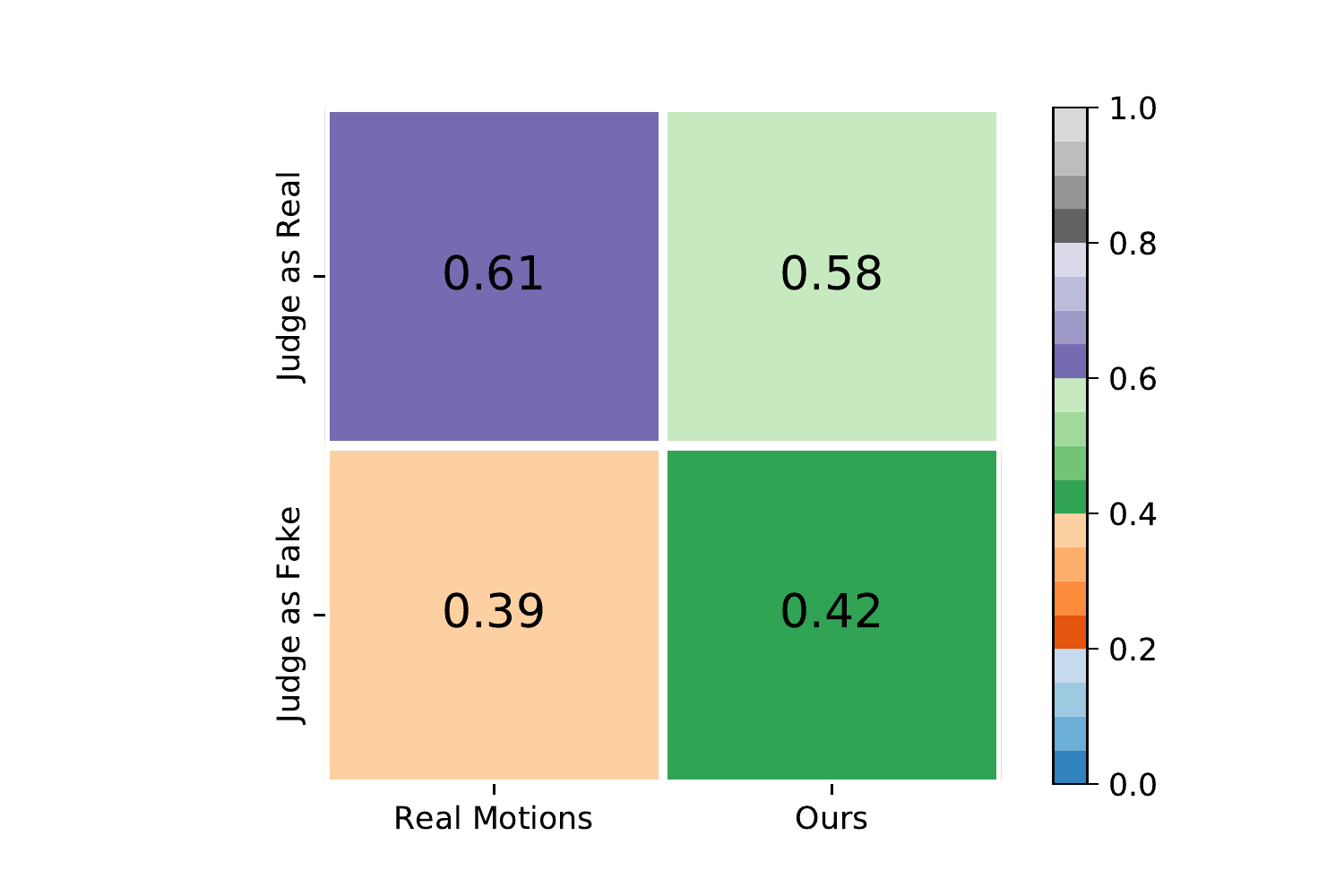}
  \vspace{-7pt}
  \setlength{\abovecaptionskip}{0.1cm}
  \setlength{\belowcaptionskip}{-0.2cm}  
  \caption{Human judge results of SMPL motions sampled from our method and real data in the testset.}
  \label{fig:real_hist}
  \vspace{-0.1cm}
\end{figure}

\subsubsection{Crowd-sourced Subjective Evaluation}
In addition to the aforementioned objective evaluation, we also conduct visual cognitive evaluation: two user studies are conducted, involving 20 subjects of various age, gender, and race. 

In the first user study, motions generated from multiple methods for the same action type are mixed together; users are then asked to rank their preference over these motions. The ranking is based on the visual perceptual quality of the generated motion, and how well it is matched with its action category. 
Fig.~\ref{fig:rank_hist} illustrates the preference results. 
Compared to comparison methods, action2motion earns most appreciation from users; on the other hand, conditional RNN is clearly the least preferred choice, with two-stage GAN being the second least; Act-MoCoGAN is relatively more preferred, but still lacks positive attention from users. Ours w/o Lie starts to generate user friendly motions, with 20\% motions are ranked at the first place by users. Action2motion further bridges the gap to real motions, with 64\% generated motions being placed at top 2 positions by users. This human study solidly substantiates the capability of our approach toward synthesizing visually pleasing motions.

To further analyze the potential of our generated motions, we render a SMPL shape for each motion and ask users to discriminate whether it is real or fake (i.e. generated) of a known action type. Specifically, the second survey consists of 72 motions (i.e. half of them are generated, and half real) uniformly sampled from across all action types. 
As shown in Fig.~\ref{fig:real_hist}, the motions from action2motion is visually only slightly inferior to real-life human motions: 
58\% of our generated motions are considered real by users, which is only 3\% lower than the number from real motions. 
The result suggests potentials of more interested downstream VR/AR applications, such as photo-realistic human video generation in gaming.

\begin{figure}[t]
  \centering
  \includegraphics[width=0.99\linewidth]{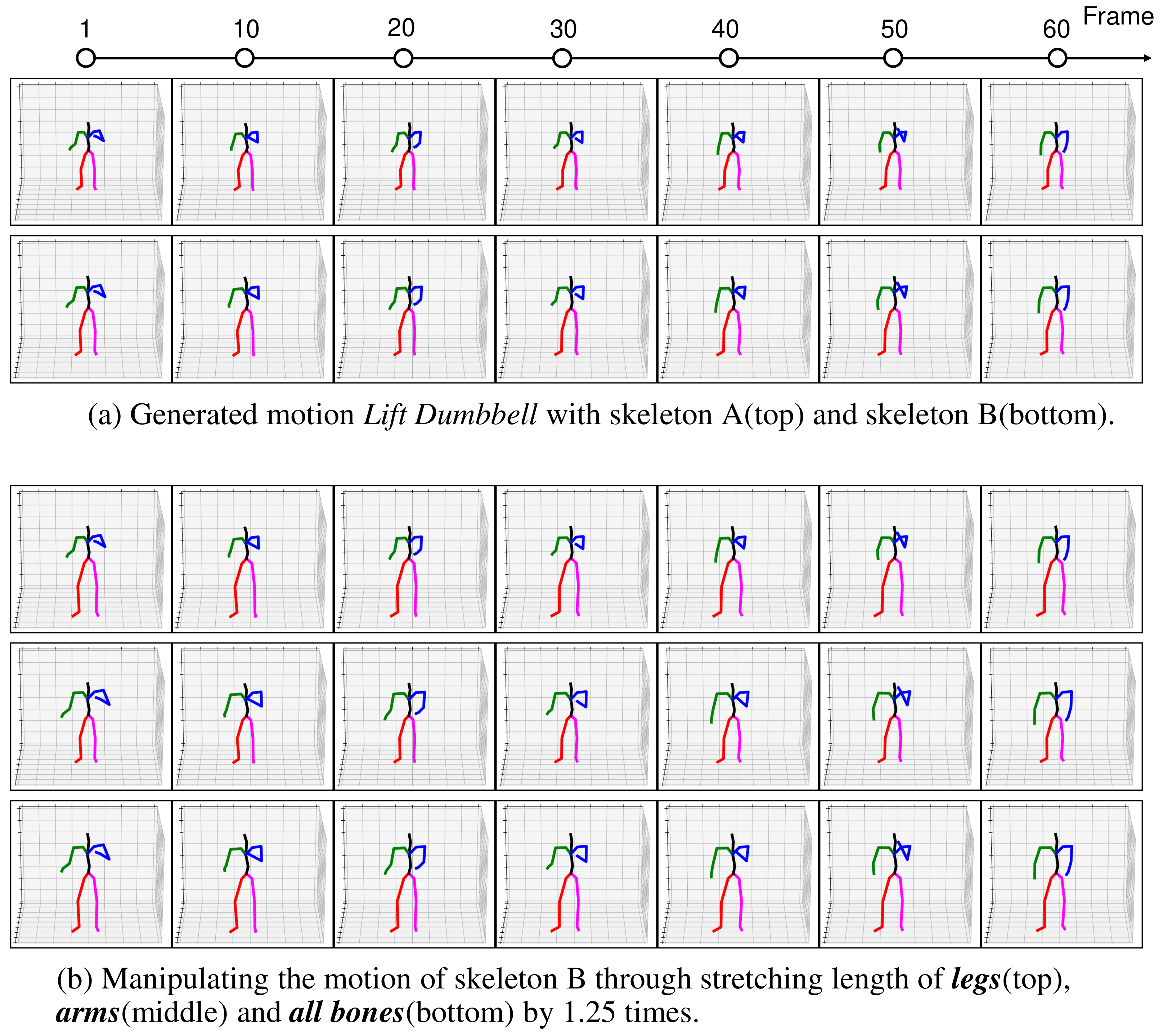}
  \setlength{\abovecaptionskip}{0.1cm}
  \setlength{\belowcaptionskip}{-0.2cm} 
  \caption{Examples of controllable motion generation. The upper sub-figure shows generating the same motion with different human skeletons. The lower sub-figure shows motions with varying scales of certain body parts.}
  \label{fig:quanlitative_result_2}
\end{figure}

\subsubsection{Controllable 3D Motion Generation}

In Figure~\ref{fig:quanlitative_result_2}, we show the capacity of our approach in generating the same 3D motions with different human skeletons. 
In action2motion, this could be easily achieved by directly modifying the bone lengths. 
In tradition methods, one could also vary the bone lengths of human motions by 3D rigid transformation. There are two advantages of our method. Firstly, it's easier to implement than transformation. Secondly, our method could change the length of arbitrary bones. 
Furthermore, our method shed new lights on disentangled human motion generation.

\section{Conclusion and Outlook}
Our paper looks at an emerging research problem of generating 3D human motions grounded on prescribed actions, and places special attention toward producing a diverse set of natural motions. 
This leads to the proposed action2motion, a conditional temporal VAE endowed with Lie algebraic representation. 
We also curate an in-house dataset and adapt two existing datasets to provide a suite of dedicated evaluation benchmarks.
Extensive qualitative, quantitative, and subjective experiments demonstrate the effectiveness of our approach against the comparison methods. 
For future work, we plan to conduct more systematic investigation on a wider range of human actions, including those involving two or more people.

\begin{acks}
This work is supported by the University of Alberta Start-up grant, the NSERC Discovery Grants including No. RGPIN-2019-04575, and the University of Alberta-Huawei Joint Innovation Collaboration grants. 
\end{acks}

\bibliographystyle{ACM-Reference-Format}
\bibliography{sample-base}


\begin{thebibliography}{39}


\ifx \showCODEN    \undefined \def \showCODEN     #1{\unskip}     \fi
\ifx \showDOI      \undefined \def \showDOI       #1{#1}\fi
\ifx \showISBNx    \undefined \def \showISBNx     #1{\unskip}     \fi
\ifx \showISBNxiii \undefined \def \showISBNxiii  #1{\unskip}     \fi
\ifx \showISSN     \undefined \def \showISSN      #1{\unskip}     \fi
\ifx \showLCCN     \undefined \def \showLCCN      #1{\unskip}     \fi
\ifx \shownote     \undefined \def \shownote      #1{#1}          \fi
\ifx \showarticletitle \undefined \def \showarticletitle #1{#1}   \fi
\ifx \showURL      \undefined \def \showURL       {\relax}        \fi
\providecommand\bibfield[2]{#2}
\providecommand\bibinfo[2]{#2}
\providecommand\natexlab[1]{#1}
\providecommand\showeprint[2][]{arXiv:#2}

\bibitem[\protect\citeauthoryear{Ahn, Ha, Choi, Yoo, and Oh}{Ahn
  et~al\mbox{.}}{2018}]%
        {ahn2018text2action}
\bibfield{author}{\bibinfo{person}{Hyemin Ahn}, \bibinfo{person}{Timothy Ha},
  \bibinfo{person}{Yunho Choi}, \bibinfo{person}{Hwiyeon Yoo}, {and}
  \bibinfo{person}{Songhwai Oh}.} \bibinfo{year}{2018}\natexlab{}.
\newblock \showarticletitle{Text2action: Generative adversarial synthesis from
  language to action}. In \bibinfo{booktitle}{\emph{Proceedings of IEEE
  International Conference on Robotics and Automation (ICRA)}}. IEEE,
  \bibinfo{pages}{5915--5920}.
\newblock


\bibitem[\protect\citeauthoryear{Ahuja and Morency}{Ahuja and Morency}{2019}]%
        {ahuja2019language2pose}
\bibfield{author}{\bibinfo{person}{Chaitanya Ahuja} {and}
  \bibinfo{person}{Louis-Philippe Morency}.} \bibinfo{year}{2019}\natexlab{}.
\newblock \showarticletitle{Language2Pose: Natural Language Grounded Pose
  Forecasting}. In \bibinfo{booktitle}{\emph{International Conference on 3D
  Vision (3DV)}}. IEEE, \bibinfo{pages}{719--728}.
\newblock


\bibitem[\protect\citeauthoryear{Cai, Bai, Tai, and Tang}{Cai
  et~al\mbox{.}}{2018}]%
        {cai2018deep}
\bibfield{author}{\bibinfo{person}{Haoye Cai}, \bibinfo{person}{Chunyan Bai},
  \bibinfo{person}{Yu-Wing Tai}, {and} \bibinfo{person}{Chi-Keung Tang}.}
  \bibinfo{year}{2018}\natexlab{}.
\newblock \showarticletitle{Deep video generation, prediction and completion of
  human action sequences}. In \bibinfo{booktitle}{\emph{Proceedings of the
  European Conference on Computer Vision (ECCV)}}. \bibinfo{pages}{366--382}.
\newblock


\bibitem[\protect\citeauthoryear{Chaaraoui, Padilla-L{\'o}pez,
  Climent-P{\'e}rez, and Fl{\'o}rez-Revuelta}{Chaaraoui et~al\mbox{.}}{2014}]%
        {chaaraoui2014evolutionary}
\bibfield{author}{\bibinfo{person}{Alexandros~Andre Chaaraoui},
  \bibinfo{person}{Jos{\'e}~Ram{\'o}n Padilla-L{\'o}pez}, \bibinfo{person}{Pau
  Climent-P{\'e}rez}, {and} \bibinfo{person}{Francisco Fl{\'o}rez-Revuelta}.}
  \bibinfo{year}{2014}\natexlab{}.
\newblock \showarticletitle{Evolutionary joint selection to improve human
  action recognition with RGB-D devices}.
\newblock \bibinfo{journal}{\emph{Expert systems with applications}}
  \bibinfo{volume}{41}, \bibinfo{number}{3} (\bibinfo{year}{2014}),
  \bibinfo{pages}{786--794}.
\newblock


\bibitem[\protect\citeauthoryear{CMU}{CMU}{2003}]%
        {cmu2003mocap}
\bibfield{author}{\bibinfo{person}{CMU}.} \bibinfo{year}{2003}\natexlab{}.
\newblock \showarticletitle{CMU graphics lab motion capture database}.
\newblock  (\bibinfo{year}{2003}).
\newblock


\bibitem[\protect\citeauthoryear{Denton and Fergus}{Denton and Fergus}{2018}]%
        {denton2018stochastic}
\bibfield{author}{\bibinfo{person}{Emily Denton} {and} \bibinfo{person}{Rob
  Fergus}.} \bibinfo{year}{2018}\natexlab{}.
\newblock \showarticletitle{Stochastic Video Generation with a Learned Prior}.
  In \bibinfo{booktitle}{\emph{International Conference on Machine Learning
  (ICML)}}. \bibinfo{pages}{1174--1183}.
\newblock


\bibitem[\protect\citeauthoryear{Gavrila, Davis, et~al\mbox{.}}{Gavrila
  et~al\mbox{.}}{1995}]%
        {gavrila1995towards}
\bibfield{author}{\bibinfo{person}{Dariu~M Gavrila}, \bibinfo{person}{Larry~S
  Davis}, {et~al\mbox{.}}} \bibinfo{year}{1995}\natexlab{}.
\newblock \showarticletitle{Towards 3-d model-based tracking and recognition of
  human movement: a multi-view approach}. In
  \bibinfo{booktitle}{\emph{International workshop on automatic face-and
  gesture-recognition}}. Citeseer, \bibinfo{pages}{272--277}.
\newblock


\bibitem[\protect\citeauthoryear{Gui, Wang, Liang, and Moura}{Gui
  et~al\mbox{.}}{2018}]%
        {gui2018adversarial}
\bibfield{author}{\bibinfo{person}{Liang-Yan Gui}, \bibinfo{person}{Yu-Xiong
  Wang}, \bibinfo{person}{Xiaodan Liang}, {and} \bibinfo{person}{Jos{\'e}~MF
  Moura}.} \bibinfo{year}{2018}\natexlab{}.
\newblock \showarticletitle{Adversarial geometry-aware human motion
  prediction}. In \bibinfo{booktitle}{\emph{Proceedings of the European
  Conference on Computer Vision (ECCV)}}. \bibinfo{pages}{786--803}.
\newblock


\bibitem[\protect\citeauthoryear{Han, Reily, Hoff, and Zhang}{Han
  et~al\mbox{.}}{2017}]%
        {han2017space}
\bibfield{author}{\bibinfo{person}{Fei Han}, \bibinfo{person}{Brian Reily},
  \bibinfo{person}{William Hoff}, {and} \bibinfo{person}{Hao Zhang}.}
  \bibinfo{year}{2017}\natexlab{}.
\newblock \showarticletitle{Space-time representation of people based on 3D
  skeletal data: A review}.
\newblock \bibinfo{journal}{\emph{Computer Vision and Image Understanding}}
  \bibinfo{volume}{158} (\bibinfo{year}{2017}), \bibinfo{pages}{85--105}.
\newblock


\bibitem[\protect\citeauthoryear{Heusel, Ramsauer, Unterthiner, Nessler, and
  Hochreiter}{Heusel et~al\mbox{.}}{2017}]%
        {heusel2017gans}
\bibfield{author}{\bibinfo{person}{Martin Heusel}, \bibinfo{person}{Hubert
  Ramsauer}, \bibinfo{person}{Thomas Unterthiner}, \bibinfo{person}{Bernhard
  Nessler}, {and} \bibinfo{person}{Sepp Hochreiter}.}
  \bibinfo{year}{2017}\natexlab{}.
\newblock \showarticletitle{Gans trained by a two time-scale update rule
  converge to a local nash equilibrium}. In \bibinfo{booktitle}{\emph{Advances
  in neural information processing systems}}. \bibinfo{pages}{6626--6637}.
\newblock


\bibitem[\protect\citeauthoryear{Huang, Wan, Probst, and Van~Gool}{Huang
  et~al\mbox{.}}{2017}]%
        {huang2017deep}
\bibfield{author}{\bibinfo{person}{Zhiwu Huang}, \bibinfo{person}{Chengde Wan},
  \bibinfo{person}{Thomas Probst}, {and} \bibinfo{person}{Luc Van~Gool}.}
  \bibinfo{year}{2017}\natexlab{}.
\newblock \showarticletitle{Deep learning on lie groups for skeleton-based
  action recognition}. In \bibinfo{booktitle}{\emph{Proceedings of the IEEE
  conference on computer vision and pattern recognition (CVPR)}}.
  \bibinfo{pages}{6099--6108}.
\newblock


\bibitem[\protect\citeauthoryear{Hussein, Torki, Gowayyed, and
  El-Saban}{Hussein et~al\mbox{.}}{2013}]%
        {hussein2013human}
\bibfield{author}{\bibinfo{person}{Mohamed~E Hussein}, \bibinfo{person}{Marwan
  Torki}, \bibinfo{person}{Mohammad~A Gowayyed}, {and} \bibinfo{person}{Motaz
  El-Saban}.} \bibinfo{year}{2013}\natexlab{}.
\newblock \showarticletitle{Human action recognition using a temporal hierarchy
  of covariance descriptors on 3d joint locations}. In
  \bibinfo{booktitle}{\emph{Twenty-Third International Joint Conference on
  Artificial Intelligence (IJCAI)}}.
\newblock


\bibitem[\protect\citeauthoryear{Kim, Nam, Cho, and Kim}{Kim
  et~al\mbox{.}}{2019}]%
        {kim2019unsupervised}
\bibfield{author}{\bibinfo{person}{Yunji Kim}, \bibinfo{person}{Seonghyeon
  Nam}, \bibinfo{person}{In Cho}, {and} \bibinfo{person}{Seon~Joo Kim}.}
  \bibinfo{year}{2019}\natexlab{}.
\newblock \showarticletitle{Unsupervised Keypoint Learning for Guiding
  Class-Conditional Video Prediction}. In \bibinfo{booktitle}{\emph{Advances in
  Neural Information Processing Systems}}. \bibinfo{pages}{3809--3819}.
\newblock


\bibitem[\protect\citeauthoryear{Kingma and Welling}{Kingma and
  Welling}{2014}]%
        {kingma2013auto}
\bibfield{author}{\bibinfo{person}{Diederik~P Kingma} {and}
  \bibinfo{person}{Max Welling}.} \bibinfo{year}{2014}\natexlab{}.
\newblock \showarticletitle{Auto-encoding variational bayes}. In
  \bibinfo{booktitle}{\emph{International Conference on Learning
  Representations (ICLR)}}.
\newblock


\bibitem[\protect\citeauthoryear{Kocabas, Athanasiou, and Black}{Kocabas
  et~al\mbox{.}}{2020}]%
        {kocabas2019vibe}
\bibfield{author}{\bibinfo{person}{Muhammed Kocabas}, \bibinfo{person}{Nikos
  Athanasiou}, {and} \bibinfo{person}{Michael~J. Black}.}
  \bibinfo{year}{2020}\natexlab{}.
\newblock \showarticletitle{VIBE: Video Inference for Human Body Pose and Shape
  Estimation}. In \bibinfo{booktitle}{\emph{Proceedings of the IEEE conference
  on computer vision and pattern recognition (CVPR)}}.
\newblock


\bibitem[\protect\citeauthoryear{Lee, Yang, Liu, Wang, Lu, Yang, and Kautz}{Lee
  et~al\mbox{.}}{2019}]%
        {lee2019dancing}
\bibfield{author}{\bibinfo{person}{Hsin-Ying Lee}, \bibinfo{person}{Xiaodong
  Yang}, \bibinfo{person}{Ming-Yu Liu}, \bibinfo{person}{Ting-Chun Wang},
  \bibinfo{person}{Yu-Ding Lu}, \bibinfo{person}{Ming-Hsuan Yang}, {and}
  \bibinfo{person}{Jan Kautz}.} \bibinfo{year}{2019}\natexlab{}.
\newblock \showarticletitle{Dancing to Music}. In
  \bibinfo{booktitle}{\emph{Advances in Neural Information Processing
  Systems}}. \bibinfo{pages}{3581--3591}.
\newblock


\bibitem[\protect\citeauthoryear{Li, Zhang, and Liu}{Li et~al\mbox{.}}{2010}]%
        {li2010action}
\bibfield{author}{\bibinfo{person}{Wanqing Li}, \bibinfo{person}{Zhengyou
  Zhang}, {and} \bibinfo{person}{Zicheng Liu}.}
  \bibinfo{year}{2010}\natexlab{}.
\newblock \showarticletitle{Action recognition based on a bag of 3d points}. In
  \bibinfo{booktitle}{\emph{2010 IEEE Computer Society Conference on Computer
  Vision and Pattern Recognition-Workshops}}. IEEE, \bibinfo{pages}{9--14}.
\newblock


\bibitem[\protect\citeauthoryear{Lin, Wu, Corona, Tai, Huang, and Mooney}{Lin
  et~al\mbox{.}}{2018}]%
        {lin20181}
\bibfield{author}{\bibinfo{person}{Angela~S Lin}, \bibinfo{person}{Lemeng Wu},
  \bibinfo{person}{Rodolfo Corona}, \bibinfo{person}{Kevin Tai},
  \bibinfo{person}{Qixing Huang}, {and} \bibinfo{person}{Raymond~J Mooney}.}
  \bibinfo{year}{2018}\natexlab{}.
\newblock \showarticletitle{generating animated videos of human activities from
  natural language descriptions}. In \bibinfo{booktitle}{\emph{Proceedings of
  the Visually Grounded Interaction and Language Workshop at NeurIPS 2018}}.
\newblock


\bibitem[\protect\citeauthoryear{Liu, Shahroudy, Perez, Wang, Duan, and
  Chichung}{Liu et~al\mbox{.}}{2019a}]%
        {liu2019ntu}
\bibfield{author}{\bibinfo{person}{Jun Liu}, \bibinfo{person}{Amir Shahroudy},
  \bibinfo{person}{Mauricio~Lisboa Perez}, \bibinfo{person}{Gang Wang},
  \bibinfo{person}{Ling-Yu Duan}, {and} \bibinfo{person}{Alex~Kot Chichung}.}
  \bibinfo{year}{2019}\natexlab{a}.
\newblock \showarticletitle{NTU RGB+D 120: A Large-Scale Benchmark for 3D Human
  Activity Understanding}.
\newblock \bibinfo{journal}{\emph{IEEE transactions on pattern analysis and
  machine intelligence}} (\bibinfo{year}{2019}).
\newblock


\bibitem[\protect\citeauthoryear{Liu, Wu, Jin, Liu, Lu, Zimmermann, and
  Cheng}{Liu et~al\mbox{.}}{2019b}]%
        {liu2019towards}
\bibfield{author}{\bibinfo{person}{Zhenguang Liu}, \bibinfo{person}{Shuang Wu},
  \bibinfo{person}{Shuyuan Jin}, \bibinfo{person}{Qi Liu},
  \bibinfo{person}{Shijian Lu}, \bibinfo{person}{Roger Zimmermann}, {and}
  \bibinfo{person}{Li Cheng}.} \bibinfo{year}{2019}\natexlab{b}.
\newblock \showarticletitle{Towards natural and accurate future motion
  prediction of humans and animals}. In \bibinfo{booktitle}{\emph{Proceedings
  of the IEEE Conference on Computer Vision and Pattern Recognition (CVPR)}}.
  \bibinfo{pages}{10004--10012}.
\newblock


\bibitem[\protect\citeauthoryear{M{\"u}ller}{M{\"u}ller}{2007}]%
        {muller2007information}
\bibfield{author}{\bibinfo{person}{Meinard M{\"u}ller}.}
  \bibinfo{year}{2007}\natexlab{}.
\newblock \bibinfo{booktitle}{\emph{Information retrieval for music and
  motion}}. Vol.~\bibinfo{volume}{2}.
\newblock \bibinfo{publisher}{Springer}.
\newblock


\bibitem[\protect\citeauthoryear{M{\"u}ller, R{\"o}der, Clausen, Eberhardt,
  Kr{\"u}ger, and Weber}{M{\"u}ller et~al\mbox{.}}{[n.d.]}]%
        {muller2007mocap}
\bibfield{author}{\bibinfo{person}{Meinard M{\"u}ller}, \bibinfo{person}{Tido
  R{\"o}der}, \bibinfo{person}{Michael Clausen}, \bibinfo{person}{Bernhard
  Eberhardt}, \bibinfo{person}{Bj{\"o}rn Kr{\"u}ger}, {and}
  \bibinfo{person}{Andreas Weber}.} \bibinfo{year}{[n.d.]}\natexlab{}.
\newblock \showarticletitle{Mocap database hdm05}.
\newblock  (\bibinfo{year}{[n.\,d.]}).
\newblock


\bibitem[\protect\citeauthoryear{Murray, Li, and Sastry}{Murray
  et~al\mbox{.}}{1994}]%
        {murray1994mathematical}
\bibfield{author}{\bibinfo{person}{Richard~M Murray}, \bibinfo{person}{Zexiang
  Li}, {and} \bibinfo{person}{S~Shankar Sastry}.}
  \bibinfo{year}{1994}\natexlab{}.
\newblock \bibinfo{booktitle}{\emph{A mathematical introduction to robotic
  manipulation}}.
\newblock \bibinfo{publisher}{CRC press}.
\newblock


\bibitem[\protect\citeauthoryear{Plappert, Mandery, and Asfour}{Plappert
  et~al\mbox{.}}{2018}]%
        {plappert2018learning}
\bibfield{author}{\bibinfo{person}{Matthias Plappert},
  \bibinfo{person}{Christian Mandery}, {and} \bibinfo{person}{Tamim Asfour}.}
  \bibinfo{year}{2018}\natexlab{}.
\newblock \showarticletitle{Learning a bidirectional mapping between human
  whole-body motion and natural language using deep recurrent neural networks}.
\newblock \bibinfo{journal}{\emph{Robotics and Autonomous Systems}}
  \bibinfo{volume}{109} (\bibinfo{year}{2018}), \bibinfo{pages}{13--26}.
\newblock


\bibitem[\protect\citeauthoryear{Shlizerman, Dery, Schoen, and
  Kemelmacher-Shlizerman}{Shlizerman et~al\mbox{.}}{2018}]%
        {shlizerman2018audio}
\bibfield{author}{\bibinfo{person}{Eli Shlizerman}, \bibinfo{person}{Lucio
  Dery}, \bibinfo{person}{Hayden Schoen}, {and} \bibinfo{person}{Ira
  Kemelmacher-Shlizerman}.} \bibinfo{year}{2018}\natexlab{}.
\newblock \showarticletitle{Audio to body dynamics}. In
  \bibinfo{booktitle}{\emph{Proceedings of the IEEE Conference on Computer
  Vision and Pattern Recognition (CVPR)}}. \bibinfo{pages}{7574--7583}.
\newblock


\bibitem[\protect\citeauthoryear{Stoll, Camgoz, Hadfield, and Bowden}{Stoll
  et~al\mbox{.}}{2020}]%
        {stoll2020text2sign}
\bibfield{author}{\bibinfo{person}{Stephanie Stoll},
  \bibinfo{person}{Necati~Cihan Camgoz}, \bibinfo{person}{Simon Hadfield},
  {and} \bibinfo{person}{Richard Bowden}.} \bibinfo{year}{2020}\natexlab{}.
\newblock \showarticletitle{Text2Sign: Towards Sign Language Production Using
  Neural Machine Translation and Generative Adversarial Networks}.
\newblock \bibinfo{journal}{\emph{International Journal of Computer Vision}}
  \bibinfo{volume}{128} (\bibinfo{year}{2020}), \bibinfo{pages}{891–908}.
\newblock


\bibitem[\protect\citeauthoryear{Suwajanakorn, Seitz, and
  Kemelmacher-Shlizerman}{Suwajanakorn et~al\mbox{.}}{2015}]%
        {suwajanakorn2015makes}
\bibfield{author}{\bibinfo{person}{Supasorn Suwajanakorn},
  \bibinfo{person}{Steven~M Seitz}, {and} \bibinfo{person}{Ira
  Kemelmacher-Shlizerman}.} \bibinfo{year}{2015}\natexlab{}.
\newblock \showarticletitle{What makes tom hanks look like tom hanks}. In
  \bibinfo{booktitle}{\emph{Proceedings of the IEEE International Conference on
  Computer Vision (ICCV)}}. \bibinfo{pages}{3952--3960}.
\newblock


\bibitem[\protect\citeauthoryear{Takeuchi, Hasegawa, Shirakawa, Kaneko, Sakuta,
  and Sumi}{Takeuchi et~al\mbox{.}}{2017}]%
        {takeuchi2017speech}
\bibfield{author}{\bibinfo{person}{Kenta Takeuchi}, \bibinfo{person}{Dai
  Hasegawa}, \bibinfo{person}{Shinichi Shirakawa}, \bibinfo{person}{Naoshi
  Kaneko}, \bibinfo{person}{Hiroshi Sakuta}, {and} \bibinfo{person}{Kazuhiko
  Sumi}.} \bibinfo{year}{2017}\natexlab{}.
\newblock \showarticletitle{Speech-to-gesture generation: A challenge in deep
  learning approach with bi-directional LSTM}. In
  \bibinfo{booktitle}{\emph{Proceedings of the 5th International Conference on
  Human Agent Interaction}}. \bibinfo{pages}{365--369}.
\newblock


\bibitem[\protect\citeauthoryear{Tang, Jia, and Mao}{Tang
  et~al\mbox{.}}{2018}]%
        {tang2018dance}
\bibfield{author}{\bibinfo{person}{Taoran Tang}, \bibinfo{person}{Jia Jia},
  {and} \bibinfo{person}{Hanyang Mao}.} \bibinfo{year}{2018}\natexlab{}.
\newblock \showarticletitle{Dance with melody: An lstm-autoencoder approach to
  music-oriented dance synthesis}. In \bibinfo{booktitle}{\emph{Proceedings of
  the 26th ACM international conference on Multimedia (ACM MM)}}.
  \bibinfo{pages}{1598--1606}.
\newblock


\bibitem[\protect\citeauthoryear{Tulyakov, Liu, Yang, and Kautz}{Tulyakov
  et~al\mbox{.}}{2018}]%
        {tulyakov2018mocogan}
\bibfield{author}{\bibinfo{person}{Sergey Tulyakov}, \bibinfo{person}{Ming-Yu
  Liu}, \bibinfo{person}{Xiaodong Yang}, {and} \bibinfo{person}{Jan Kautz}.}
  \bibinfo{year}{2018}\natexlab{}.
\newblock \showarticletitle{Mocogan: Decomposing motion and content for video
  generation}. In \bibinfo{booktitle}{\emph{Proceedings of the IEEE conference
  on computer vision and pattern recognition (CVPR)}}.
  \bibinfo{pages}{1526--1535}.
\newblock


\bibitem[\protect\citeauthoryear{Vemulapalli, Arrate, and
  Chellappa}{Vemulapalli et~al\mbox{.}}{2014}]%
        {vemulapalli2014human}
\bibfield{author}{\bibinfo{person}{Raviteja Vemulapalli},
  \bibinfo{person}{Felipe Arrate}, {and} \bibinfo{person}{Rama Chellappa}.}
  \bibinfo{year}{2014}\natexlab{}.
\newblock \showarticletitle{Human action recognition by representing 3d
  skeletons as points in a lie group}. In \bibinfo{booktitle}{\emph{Proceedings
  of the IEEE conference on computer vision and pattern recognition (CVPR)}}.
  \bibinfo{pages}{588--595}.
\newblock


\bibitem[\protect\citeauthoryear{Wang, Liu, Wu, and Yuan}{Wang
  et~al\mbox{.}}{2012}]%
        {wang2012mining}
\bibfield{author}{\bibinfo{person}{Jiang Wang}, \bibinfo{person}{Zicheng Liu},
  \bibinfo{person}{Ying Wu}, {and} \bibinfo{person}{Junsong Yuan}.}
  \bibinfo{year}{2012}\natexlab{}.
\newblock \showarticletitle{Mining actionlet ensemble for action recognition
  with depth cameras}. In \bibinfo{booktitle}{\emph{IEEE Conference on Computer
  Vision and Pattern Recognition (CVPR)}}. IEEE, \bibinfo{pages}{1290--1297}.
\newblock


\bibitem[\protect\citeauthoryear{Xia, Chen, and Aggarwal}{Xia
  et~al\mbox{.}}{2012}]%
        {xia2012view}
\bibfield{author}{\bibinfo{person}{Lu Xia}, \bibinfo{person}{Chia-Chih Chen},
  {and} \bibinfo{person}{Jake~K Aggarwal}.} \bibinfo{year}{2012}\natexlab{}.
\newblock \showarticletitle{View invariant human action recognition using
  histograms of 3d joints}. In \bibinfo{booktitle}{\emph{2012 IEEE Computer
  Society Conference on Computer Vision and Pattern Recognition Workshops}}.
  IEEE, \bibinfo{pages}{20--27}.
\newblock


\bibitem[\protect\citeauthoryear{Xu, Govindarajan, Zhang, and Cheng}{Xu
  et~al\mbox{.}}{2017}]%
        {xu2017lie}
\bibfield{author}{\bibinfo{person}{Chi Xu}, \bibinfo{person}{Lakshmi~Narasimhan
  Govindarajan}, \bibinfo{person}{Yu Zhang}, {and} \bibinfo{person}{Li Cheng}.}
  \bibinfo{year}{2017}\natexlab{}.
\newblock \showarticletitle{Lie-X: Depth image based articulated object pose
  estimation, tracking, and action recognition on lie groups}.
\newblock \bibinfo{journal}{\emph{International Journal of Computer Vision}}
  \bibinfo{volume}{123}, \bibinfo{number}{3} (\bibinfo{year}{2017}),
  \bibinfo{pages}{454--478}.
\newblock


\bibitem[\protect\citeauthoryear{Yacoob and Black}{Yacoob and Black}{1999}]%
        {yacoob1999parameterized}
\bibfield{author}{\bibinfo{person}{Yaser Yacoob} {and}
  \bibinfo{person}{Michael~J Black}.} \bibinfo{year}{1999}\natexlab{}.
\newblock \showarticletitle{Parameterized modeling and recognition of
  activities}.
\newblock \bibinfo{journal}{\emph{Computer Vision and Image Understanding}}
  \bibinfo{volume}{73}, \bibinfo{number}{2} (\bibinfo{year}{1999}),
  \bibinfo{pages}{232--247}.
\newblock


\bibitem[\protect\citeauthoryear{Yan, Rastogi, Villegas, Sunkavalli, Shechtman,
  Hadap, Yumer, and Lee}{Yan et~al\mbox{.}}{2018}]%
        {yan2018mt}
\bibfield{author}{\bibinfo{person}{Xinchen Yan}, \bibinfo{person}{Akash
  Rastogi}, \bibinfo{person}{Ruben Villegas}, \bibinfo{person}{Kalyan
  Sunkavalli}, \bibinfo{person}{Eli Shechtman}, \bibinfo{person}{Sunil Hadap},
  \bibinfo{person}{Ersin Yumer}, {and} \bibinfo{person}{Honglak Lee}.}
  \bibinfo{year}{2018}\natexlab{}.
\newblock \showarticletitle{Mt-vae: Learning motion transformations to generate
  multimodal human dynamics}. In \bibinfo{booktitle}{\emph{Proceedings of the
  European Conference on Computer Vision (ECCV)}}. \bibinfo{pages}{265--281}.
\newblock


\bibitem[\protect\citeauthoryear{Yang, Wang, Zhu, Huang, Shi, and Lin}{Yang
  et~al\mbox{.}}{2018}]%
        {yang2018pose}
\bibfield{author}{\bibinfo{person}{Ceyuan Yang}, \bibinfo{person}{Zhe Wang},
  \bibinfo{person}{Xinge Zhu}, \bibinfo{person}{Chen Huang},
  \bibinfo{person}{Jianping Shi}, {and} \bibinfo{person}{Dahua Lin}.}
  \bibinfo{year}{2018}\natexlab{}.
\newblock \showarticletitle{Pose guided human video generation}. In
  \bibinfo{booktitle}{\emph{Proceedings of the European Conference on Computer
  Vision (ECCV)}}. \bibinfo{pages}{201--216}.
\newblock


\bibitem[\protect\citeauthoryear{Zou, Zuo, Qian, Wang, Xu, Gong, and Cheng}{Zou
  et~al\mbox{.}}{2020a}]%
        {zou2020detailed}
\bibfield{author}{\bibinfo{person}{Shihao Zou}, \bibinfo{person}{Xinxin Zuo},
  \bibinfo{person}{Yiming Qian}, \bibinfo{person}{Sen Wang},
  \bibinfo{person}{Chi Xu}, \bibinfo{person}{Minglun Gong}, {and}
  \bibinfo{person}{Li Cheng}.} \bibinfo{year}{2020}\natexlab{a}.
\newblock \showarticletitle{3D Human Shape Reconstruction from a Polarization
  Image}. In \bibinfo{booktitle}{\emph{Proceedings of the European Conference
  on Computer Vision (ECCV)}}.
\newblock


\bibitem[\protect\citeauthoryear{Zou, Zuo, Qian, Wang, Xu, Gong, and Cheng}{Zou
  et~al\mbox{.}}{2020b}]%
        {polardataset}
\bibfield{author}{\bibinfo{person}{Shihao Zou}, \bibinfo{person}{Xinxin Zuo},
  \bibinfo{person}{Yiming Qian}, \bibinfo{person}{Sen Wang},
  \bibinfo{person}{Chi Xu}, \bibinfo{person}{Minglun Gong}, {and}
  \bibinfo{person}{Li Cheng}.} \bibinfo{year}{2020}\natexlab{b}.
\newblock \bibinfo{title}{Polarization Human Shape and Pose Dataset}.
\newblock
\newblock
\showeprint{arXiv:2004.14899}


\end{thebibliography}



\appendix

\section{Appendix}

\subsection{Neural Network Architecture}

\begin{table*}[htbp]
  \caption{Architecture of our action2motion model on dataset \DN.}
  \label{tab:architecure}
  \begin{tabular}{l c c}
    \toprule
                Components & Architecture & Number of Params \\
    \midrule
    \multirow{4}{*}{Posterior Network} & \multicolumn{1}{l}{(encoder): Linear(in\_features=85, out\_features=128, bias=True)} & \multirow{4}{*}{117,820}\\
    & (gru): ModuleList((0): GRUCell(128, 128)) \\
    & (mu\_net): Linear(in\_features=128, out\_features=30, bias=True)\\
    & (logvar\_net): Linear(in\_features=128, out\_features=30, bias=True)\\
    \midrule
    \multirow{4}{*}{Prior Network} & \multicolumn{1}{l}{(encoder): Linear(in\_features=85, out\_features=128, bias=True)} &\multirow{4}{*}{117,820}\\
    & (gru): ModuleList((0): GRUCell(128, 128)) \\
    & (mu\_net): Linear(in\_features=128, out\_features=30, bias=True)\\
    & (logvar\_net): Linear(in\_features=128, out\_features=30, bias=True)\\
    \midrule
    \multirow{4}{*}{Generator Network} & \multicolumn{1}{l}{(encoder): Linear(in\_features=115, out\_features=128, bias=True))} &\multirow{4}{*}{227,110}\\
    & (gru): ModuleList(
    (0): GRUCell(128, 128)
    (1): GRUCell(128, 128)
  ) \\
    & (decoder): Linear(in\_features=128, out\_features=72, bias=True)\\
    & (Lie\_output): Linear(in\_features=69, out\_features=69, bias=True)\\
    \bottomrule
  \end{tabular}
\end{table*}

Table \ref{tab:architecure} illustrates the architecture we used on dataset \DN. For other two datasets, the dimension of input vector may vary according to the dimension of pose vector.

\subsection{Implementation Details}

Our approach is implemented by PyTorch. The output size of all encoder layers and decoder layer is set to 128 and 72, respectively. A two-layer GRU is used for generator, and one layer GRU is for both prior network and posterior network, all having the same hidden unit size of 128. The noise vector $\mathbf{z}$ is 30 dimensional. Altogether, our learned model has about 450,000 parameters. The Adam optimizer is applied for training in all experiments, with learning rate of 0.0002, weight decaying of 0.00001, and default parameter values including $\beta_1=0.9$, $\beta_2=0.999$. We train our model with a mini-batch size of 128. To stabilize the training, the \textit{teacher forcing rate} $p_{\mathrm{tf}}$ is set to 0.6. Values of the above hyper-parameters are fixed throughout empirical evaluations across all datasets. 
Afterwards, we generate motion with length of 60, 100 and 60 on NTU-RGB-D, CMU MoCap and \DN, respectively. The hyper-parameter $\lambda$ is a trade-off between reconstruction constraints and KL-divergence penalty, with its value in the above datasets setting to 0.1, 0.1 and 0.01, respectively. Empirically, a large lambda will increase the quality of generated motion but may decrease motion diversity; a flipped effect is held for a small lambda.

\begin{figure}[htb]
  \centering
  \includegraphics[width=0.8\linewidth]{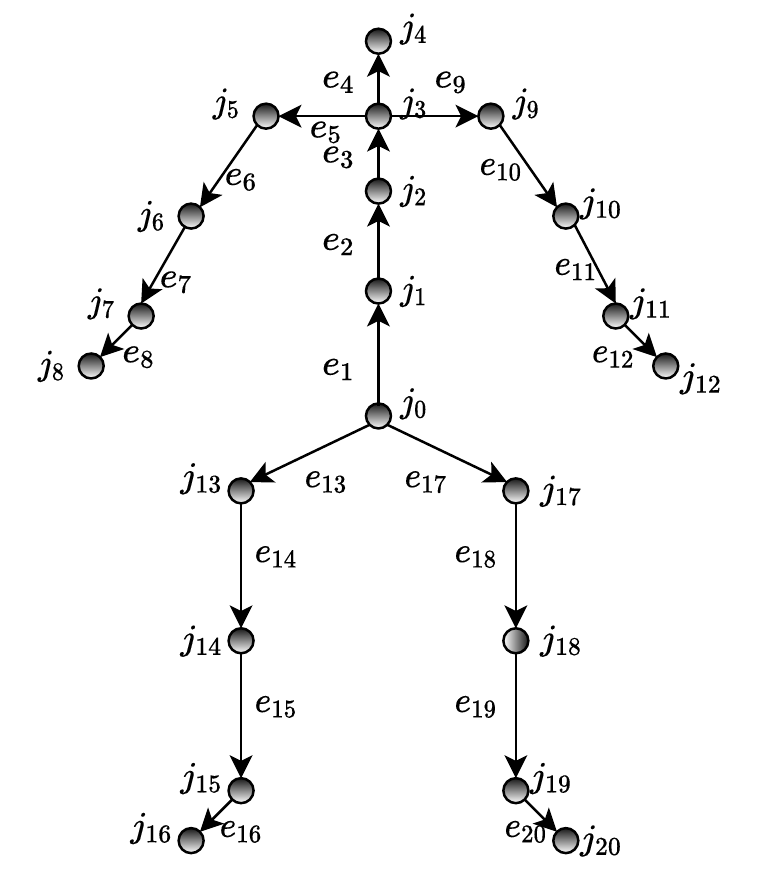}
  \caption{An example of human skeleton consisting of 21 joints and 20 body parts.}
  \label{fig:skeleton}
\end{figure}

\subsection{Exemplar Human Skeleton}
Fig.~\ref{fig:skeleton} gives an exemplar of human skeleton anatomical structure. This skeleton contains 21 joints and 20 bones, and there are five kinematic chains in this skeleton which are
\begin{itemize}
    \item spine: $[j_0, j_1, j_2, j_3, j_4]$, 
    \item left arm: $[j_3, j_5, j_6, j_7, j_8]$, 
    \item right arm: $[j_3, j_9, j_{10}, j_{11}, j_{12}]$, 
    \item left leg: $[j_0, j_{13}, j_{14}, j_{15}, j_{16}]$
    \item right leg: $[j_0, j_{17}, j_{18}, j_{19}, j_{20}]$
\end{itemize}

\subsection{Training Process Comparison}
\begin{figure*}
    \centering
    \includegraphics[width=1.05\linewidth]{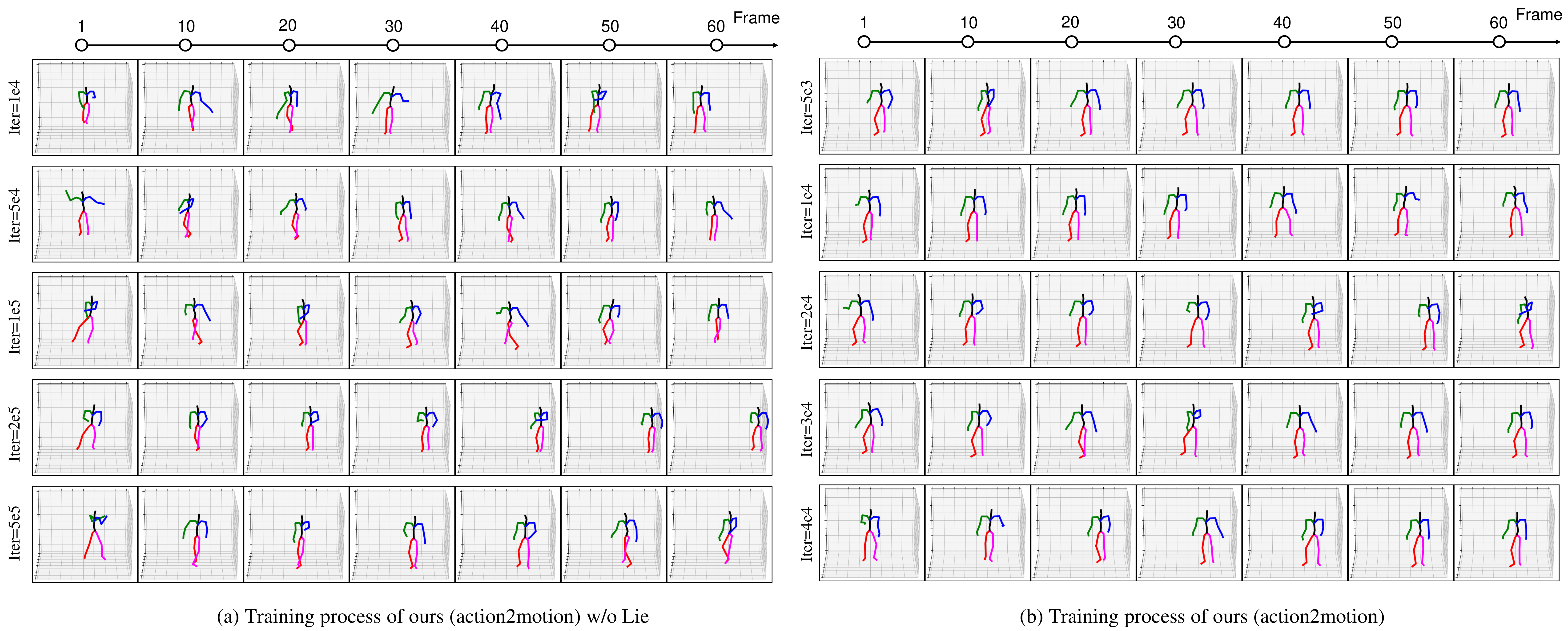}
    \caption{Generated \textit{walk} motions at different training iterations from (a) our model without Lie and (b) our model.}
    \label{fig:training processss}
\end{figure*}

Fig.\ref{fig:training processss} gives an illustrative example of the training process comparison between our model with and without Lie algebra representation. Without Lie algebra representation, our model keep generating poses with obvious artifacts at iteration 10,000, and evolves to produce natural and stable \textit{walk} motion within 500,000 iterations. Incorporation of Lie significantly cut down the number of iterations to 40,000. In addition, ours model with Lie algebra is ready to generate plausible pose sequences within 5,000 iterations.

\begin{figure*}[ht]
    \centering
    \includegraphics[width=0.95\linewidth]{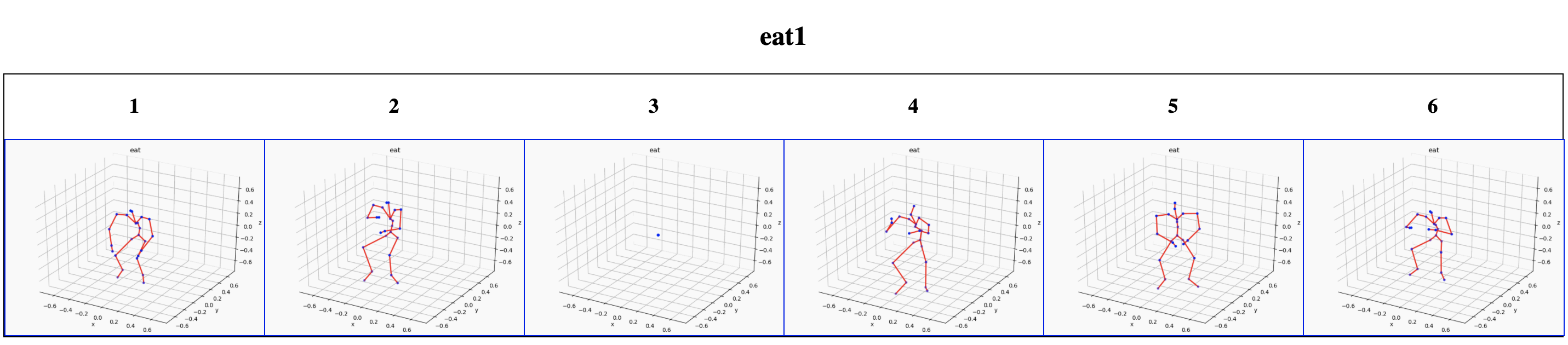}
    \includegraphics[width=0.95\linewidth]{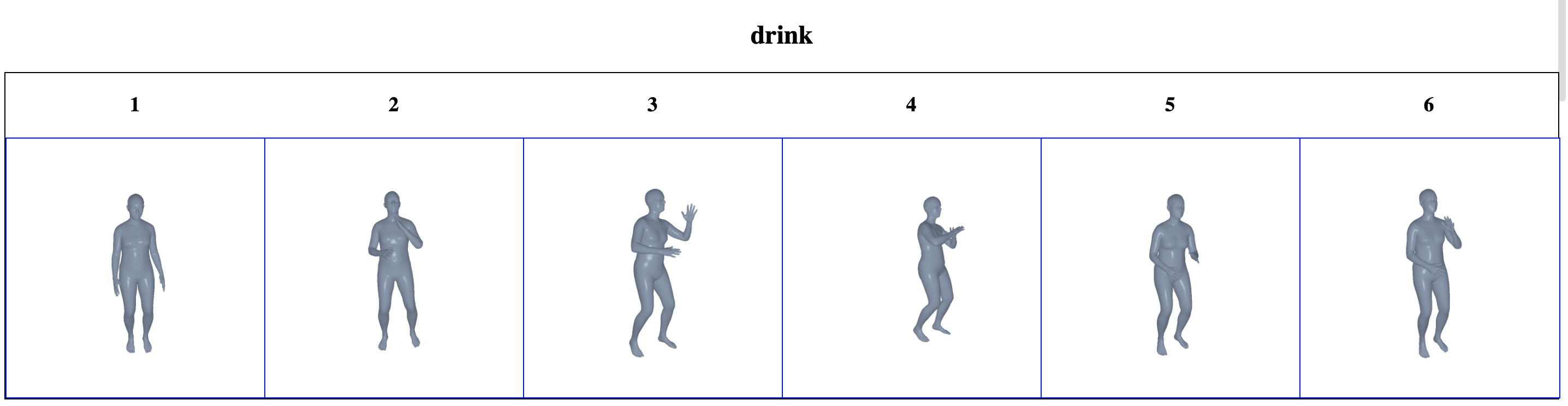}
    \caption{Examples of our two user studies: \textit{user preference}(top) and \textit{fake or real}(bottom).}
    \label{fig:user study}
\end{figure*}

\subsection{Design of User Study}
We conducted two user studies in our experiments, one is \textit{user preference}(Fig.\ref{fig:user study}(top)) survey and another is \textit{fake or real}(Fig.\ref{fig:user study}(bottom)). Practically, HTML is utilized to perform our survey for better user experience. 

In \textit{user preference} survey, for each question, motions from different sources are mixed up with random order; 6 motion clips and the action type(e.g. eat) are shown to users; and users are asked to give their preference over these 6 motion clips. In \textit{fake or real} survey, for each action type 6 motions sampled from our model and real data(not necessarily equal distributed) are placed together; and users need to discriminate each motion as the generated or the real. Users are told to try not to make decision by comparing one with another. 

\subsection{Dataset Details}

\subsubsection{Improved NTU-RGB-D}
In our experiments, the recent video 3d shape estimation method\cite{kocabas2019vibe} is employed for re-annotating partial NTU-RGB-D dataset. As shown in Tab.\ref{tab:ntu-rgb-d}, 13 action types are picked from the original dataset, forming an refined dataset with 3902 motion clips in total. We evidence the feasibility of improving NTU-RGB-D dataset for motion generation task, and may further re-annotate larger proportion of the dataset in the future.

\begin{table}[h]
  \caption{Statistics of improved NTU-RGB-D dataset.}
  \label{tab:ntu-rgb-d}
  \begin{tabular}{l c}
    \toprule
                Action Label & Number of Motions \\
    \midrule
            Butt kicks (kick backward) & 287\\
            Cheer up & 309 \\
            Hand waving & 309 \\
            Kicking something & 306 \\
            Pick up & 306 \\
            Running on the spot & 288 \\
            Salute & 308 \\
            Shake fist & 287 \\
            Side kick & 288 \\
            Sitting down & 306 \\
            Squat down & 306 \\
            Standing up (from sitting position) & 311 \\
            Throw & 305 \\
    \midrule
        \textbf{Entire Dataset} & \textbf{3902}\\
    \bottomrule
  \end{tabular}
\end{table}

\subsubsection{\DN}
Our dataset \DN is adopted from the dataset \textit{\DNT} proposed in \cite{polardataset}. The original dataset contains more dedicated multimodal human pose resources including polar images, RGB images, depth images, ultra images and 3d annotations. We select a subset of dataset \textit{\DNT}, and cut the long pose sequences into smaller pieces which aligns with prescribed action types. The statistics of our dataset is given in Tab.\ref{tab:\DN}, where there are 1061 motion clips which are categorized into 12 action classes and 34 sub-classes. Our dataset only contains motions with 3D positions and corresponding action type annotation. But others modality(i.e, RGB images, etc.) are accessible in \textit{\DNT} via the alignment between our dataset and  \textit{\DNT}.

\begin{table*}[ht]
  \caption{Statistics of dataset \DN.}
  \label{tab:\DN}
  \begin{tabular}{l l c c}
    \toprule
                Coarse-grained Label & Fine-grained Label & Number of Motions & Total Number \\
    \midrule
    \multirow{7}{*}{Warm up} & Warm\_up\_wristankle & 25 & \multirow{7}{*}{215} \\
                & Warm\_up\_pectoral & 49 \\
                & Warm\_up\_eblowback & 43 \\
                & Warm\_up\_bodylean\_right\_arm & 26\\
                & Warm\_up\_bodylean\_left\_arm & 24\\
                & Warm\_up\_bow\_right & 24 \\
                & Warm\_up\_bow\_left & 24 \\
    \midrule
    \multirow{1}{*}{Walk} & Walk & 47 & \multirow{1}{*}{47} \\
    \midrule
    \multirow{1}{*}{Run} & Run & 50 & \multirow{1}{*}{50} \\
    \midrule
    \multirow{2}{*}{Jump} & Jump\_handsup & 54 & \multirow{2}{*}{94} \\
                & Jump\_vertical & 40\\
    \midrule
    \multirow{5}{*}{Drink} & Drink\_bottle\_righthand & 27 & \multirow{5}{*}{88} \\
                & Drink\_bottle\_lefthand & 43 \\ 
                & Drink\_cup\_righthand & 11 \\
                & Drink\_cup\_lefthand & 3\\
                & Drink\_both\_hands & 4 \\
    \midrule
    \multirow{5}{*}{Lift\_dumbbell} & Lift\_dumbbell\_righthand & 45 & \multirow{5}{*}{218} \\
                & Lift\_dumbbell\_lefthand & 45 \\ 
                & Lift\_dumbbell\_bothhands & 47 \\
                & Lift\_dumbbell\_overhead & 43\\
                & Lift\_dumbbell\_bothhands\_bend\_legs & 38 \\
    \midrule
    \multirow{1}{*}{Sit} & Sit & 54 & \multirow{1}{*}{54} \\
    \midrule
    \multirow{3}{*}{Eat} & Eat\_righthand & 33 & \multirow{3}{*}{77} \\
                & Eat\_lefthand & 25 \\ 
                & Eat\_pie/burger & 19 \\
    \midrule
    \multirow{1}{*}{Turn\_steering\_wheel} & Turn\_steering\_wheel & 56 & \multirow{1}{*}{56} \\
    \midrule
    \multirow{2}{*}{Phone} & Take out phone, call and put back & 28 & \multirow{2}{*}{61} \\
                & Call with left hand & 33 \\ 
    \midrule
    \multirow{4}{*}{Boxing} & Boxing\_left\_right & 26 & \multirow{4}{*}{140} \\
                & Boxing\_left\_upwards & 39 \\
                & Boxing\_right\_upwards & 41 \\
                & Boxing\_right\_left & 34 \\
    \midrule
    \multirow{2}{*}{Throw} & Throw\_right\_hand & 53 & \multirow{2}{*}{91} \\
                & Throw\_both\_hand & 38 \\
    \midrule
    \multirow{1}{*}{\textbf{Entire Dataset}} & - & - & \multirow{1}{*}{\textbf{1191}} \\
    \bottomrule
  \end{tabular}
\end{table*}

\end{document}